\documentclass{article}

\usepackage{iclr2024_conference,times}
\usepackage[page]{appendix}

\usepackage[utf8]{inputenc} \usepackage[T1]{fontenc}    \usepackage{hyperref}       \usepackage{url}            \usepackage{booktabs}       \usepackage{amsfonts}       \usepackage{nicefrac}       \usepackage{microtype}      \usepackage{wrapfig}

\usepackage{wrapfig}
\usepackage{color, colortbl}

\definecolor{Gray}{gray}{0.9}	
\definecolor{LightCyan}{rgb}{0.88,1,1}

\usepackage[tbtags]{amsmath}
\usepackage{multicol}
\usepackage{tikz}
\usetikzlibrary{calc}
\usepackage{booktabs}
\newcommand{\ra}[1]{\renewcommand{\arraystretch}{#1}}
\allowdisplaybreaks
\usepackage{amssymb,mathrsfs}
\usepackage{amsfonts}
\usepackage{url}
\usepackage{amsthm}
\usepackage{amssymb,dsfont,amsmath,amsthm,mathtools}
\usepackage{enumitem}
\usepackage{bbm}
\usepackage{xcolor}

\usepackage{algorithmic}
\usepackage{algorithm}
\usepackage{dcolumn}
\usepackage{color, colortbl}
\definecolor{darkgreen}{RGB}{0,128,0}
\definecolor{darkorange}{RGB}{255,140,0}
\definecolor{darkblue}{RGB}{0,0,139}

\newtheorem{theorem}{Theorem}

\newtheorem{property}{Property}

\newtheorem{definition}{Definition}

\definecolor{Gray}{gray}{0.9}
\definecolor{LightCyan}{rgb}{0.88,1,1}
\definecolor{LightGreen}{rgb}{1,0.88,1}
\usepackage{xspace}
\newcolumntype{d}[1]{D{.}{.}{#1}}

\def\R{\mathbb{R}}
\def\1{\mathbbm{1}}

\def\a{\mathbf{a}}
\def\m{\mathbf{m}}
\def\b{\mathbf{b}}

\def\P{\mathbf{P}}

\def\C{\mathbf{C}}

\newcommand{\Ex}{{\rm I\kern-.3em E}}

\newcommand{\argmin}{\mathop{\mathrm{arg\,min}}}

\newcommand{\ie}{{\em i.e.,~}}

\newcommand{\ours}{\textcolor{magenta}{\texttt{FedWaD}}\xspace}

\usepackage[colorinlistoftodos,bordercolor=orange,backgroundcolor=orange!20,linecolor=orange,textsize=scriptsize]{todonotes}

\usepackage{cleveref}

\title{Federated Wasserstein Distance}

\author{Alain Rakotomamonjy \\
Criteo AI Lab\\
Paris, France\\
alain.rakoto@insa-rouen.fr \\
\And
Kimia Nadjahi\\
Computer Science and Artificial Intelligence Lab\\
MIT \\
knadjahi@mit.edu
\And
Liva Ralaivola \\
Criteo AI Lab\\
Paris, France\\
l.ralaivola@criteo.com
}

\iclrfinalcopy

\begin{document}

\maketitle

\begin{abstract}
We introduce a principled way of computing the Wasserstein distance between two distributions in a federated manner.
Namely, we show how to estimate the Wasserstein distance between two samples stored and
 kept on different devices/clients whilst a central entity/server orchestrates the computations 
 (again, without having access to the samples). To achieve this feat, we take advantage of the geometric 
properties of the Wasserstein distance -- in particular, the triangle inequality -- 
 and that of the associated {\em geodesics}: our algorithm, \ours (for Federated Wasserstein Distance), iteratively approximates 
 the Wasserstein distance by manipulating and exchanging distributions from the
  space of geodesics in lieu of the input samples. 
  In addition to establishing the convergence properties of \ours,
   we provide empirical results on federated coresets and federate 
   optimal transport dataset distance, that we respectively exploit for
   building a novel federated model and for boosting performance of popular federated learning algorithms.
\end{abstract}

 \section{Introduction}

\paragraph{Context.} Federated Learning (FL) is a form of distributed machine learning (ML) dedicated to 
train a global model from data stored on local devices/clients, while ensuring these clients never share their data 
\citep{KairouzFL,FLreview2021}. FL provides elegant and convenient solutions to concerns 
in data privacy, computational and storage costs of centralized training, and makes it
possible to take advantage of large amounts of data stored on local devices. 
A typical FL approach to learn a parameterized global model is to alternate
between the two following steps: i)
update local versions of the global model using local data, and ii) send and aggregate the 
parameters of the local models on a central server \citep{mcmahan2017communication} to update the global model.

\paragraph{Problem.} {In some practical situations, the goal is not to learn a
 prediction model,  but rather to compute a certain quantity from the data stored on
  the clients. For instance, one's goal may be to compute, in a federated way,
   some prototypes of client's data,
  that can be leveraged for federated clustering  or 
for classification models \citep{gribonval2021sketching,phillips2016coresets,munteanu2018coresets,agarwal2005geometric}. 
In another learning scenarios where data are scarce, one may
 want to look for similarity between datasets in order to evaluate dataset heterogeneity
 over clients and leverage on this information to improve the performance of
  federated learning algorithms.} 
  In this work, we address the problem of 
  computing, in a federated way, the Wasserstein distance between two distributions $\mu$ and $\nu$ when
samples from each distribution are stored on local devices. 
{A solution to this problem will be useful in the aforementioned situations, 
where the Wasserstein distance is used as a similarity measure between two datasets
and is the key tool for computing some coresets of the data distribution or cluster prototypes.} 
We provide a solution to this problem which hinges on the geometry of 
the Wasserstein distance and more specifically, its geodesics. We leverage the property
that for any element $\xi^\star$ of the geodesic between two distributions $\mu$ and $\nu$,
 the following equality holds, $\mathcal{W}_p(\mu,\nu)=\mathcal{W}_p(\mu,\xi^\star)+\mathcal{W}_p(\xi^\star,\nu)$, where
$\mathcal{W}_p$ denotes the $p$-Wasserstein distance.
This property is especially useful to compute $\mathcal{W}_p(\mu, \nu)$ in a federated manner,
 leading to a novel theoretically-justified procedure coined \ours, for {\bf Fed}erated {\bf Wa}sserstein {\bf D}istance.

\paragraph{Contribution: \ours.}
The principle of \ours is to iteratively approximate $\xi^\star$ -- which, in terms
of traditional FL, can be interpreted as the global model. At iteration $k$, our procedure 
consists in i) computing, on the clients, distributions $\xi_{\mu}^{k}$ and $\xi_{\nu}^{k}$ from 
the geodesics between the current approximation of $\xi^\star$ and the two secluded 
distributions $\mu$ and $\nu$ -- $\xi_{\mu}^{k}$ and $\xi_{\nu}^{k}$ playing the 
role of the local versions of the global model, and ii) aggregating
 them on the global model to update $\xi^\star$.

\paragraph{Organization of the paper.} \Cref{sec:background} formalizes the problem we address,
 and provides the necessary technical background to devise our algorithm~\ours. 
\Cref{sec:contribution}
is devoted to the depiction of \ours, pathways to speed-up its executions, and a theoretical justification that \ours is guaranteed to converge to the desired quantity. In \Cref{sec:results}, we conduct an empirical analysis of \ours on different use-cases 
(Wasserstein coresets and Optimal Transport Dataset distance) which rely on
 the computation of the Wasserstein distance. 
 We unveil how these problems can be solved in our FL setting and 
  demonstrates the remarkable versatility  of our approach. 
  In particular, we expose the  impact of 
  federated coresets. By  learning 
  a single global model on the server based on the coreset, our method
   can outperform personalized FL models. In addition, our ability to compute 
   inter-device dataset distances significantly helps amplify 
performances of popular federated learning algorithms, 
such as FedAvg, FedRep, and FedPer.
We achieve this by  clustering clients and 
harnessing the power of reduced dataset heterogeneity.

 \section{Related Works and Background}
 \label{sec:background}
 
\subsection{Wasserstein Distance and Geodesics}
Throughout, we denote by $\mathscr{P}(X)$ the set of probability measures in $X$. Let $p \geq 1$ and define $\mathscr{P}_p(X)$ the subset of measures in $\mathscr{P}(X)$ with finite $p$-moment, \ie
$\mathscr{P}_p(X) \doteq \big\{\eta \in \mathscr{P}(X): M_p(\eta) < \infty\big\}$, where $M_p(\eta)\doteq\int_{X} {d^p_X(x, 0)} d\eta(x)$ and $d_X$ is a metric on $X$ often referred to as
the {\em ground cost}. For $\mu \in \mathscr{P}_p(X)$ and $\nu \in \mathscr{P}_p(Y)$, 
 $\Pi(\mu, \nu) \subset \mathscr{P}(X \times Y)$ is the 
 collection of probability measures or {\em couplings} on $X \times Y$ defined as 
\begin{align*}
	\Pi(\mu, \nu) \doteq \big\{&\pi\in \mathscr{P}(X\times Y):
	 \forall A \subset X, B \subset Y, 
	 \pi(A \times Y) = \mu(A) \text{ and } \pi(X \times B) = \nu(B)\big\}.
\end{align*}

The $p$-Wasserstein distance $\mathcal{W}_p(\mu, \nu)$ between the measures $\mu$ and $\nu$ ---assumed to be defined over the same ground space, i.e. $X=Y$--- is defined as
\begin{equation}
\label{wasserstein-dist}
\mathcal{W}_p(\mu, \nu) \doteq \left ( \inf_{\pi \in \Pi(\mu, \nu)}\int_{X\times X}d^p_X(x, x')d\pi(x,x') \right)^{1/p}.
\end{equation}
It is proven that the infimum in~\eqref{wasserstein-dist} is attained \citep{peyre2019computational}
 and any probability $\pi$ which realizes 
the minimum is an {\it optimal transport plan.}
In the discrete case, we denote the
two marginal measures as $\mu = \sum_{i=1}^n a_i \delta_{x_i}$ and 
$\nu = \sum_{i=1}^m b_i \delta_{x_i^\prime}$, with $a_i, b_i \geq 0$ and $\sum_{i=1}^n a_i = \sum_{i=1}^m b_i = 1$.
 The 
{\em Kantorovitch relaxation}
of~\eqref{wasserstein-dist} seeks for a transportation coupling $\P$ that solves the problem 
\begin{equation}\label{eq:wd}
    \mathcal{W}_p(\mu, \nu)  \doteq \left ( \min_{\P \in \Pi(\a,\b)} \langle \C, \P \rangle \right)^{1/p}
\end{equation}
where $\C\doteq(d_X^p(x_i,x_j^\prime)) \in \R^{n \times m}$ is the matrix of all pairwise
costs, and 
$\Pi(\a,\b) \doteq \{\P \in \R_+^{n\times m}| \P \mathbf{1} = \a, \P^\top \mathbf{1} = \b \}$ is the {\em transportation polytope} (i.e. the set of all transportation plans) between the distributions $\a$ and $\b$. 

\begin{property}[\cite{peyre2019computational}]
For any $p\geq 1$, $\mathcal{W}_p$ {\em is} a metric on $\mathscr{P}_p(X)$. As such it satisfies
the triangle inequality:
\begin{equation}
	\label{eq:triangle}
	\forall \mu,\nu,\xi\in\mathscr{P}_p(X),\quad \mathcal{W}_p(\mu,\nu)\leq \mathcal{W}_p(\mu,\xi) + \mathcal{W}_p(\xi, \nu)
\end{equation}
\end{property}

It might be convenient to consider {\em geodesics} as structuring tools of metric spaces.
\begin{definition}[Geodesics, \citep{ambrosio2005gradient}]
Let ($\mathcal{X}, \text{d}$) be a metric space. A {\em constant speed geodesic} $x : [0, 1] \rightarrow \mathcal{X}$ between $x_0,x_1$ $\in  \mathcal{X}$ is a continuous curve 
such that $\forall s, t \in [0, 1]$, $\text{d}(x(s), x(t)) = |s - t| \cdot\text{d}(x_0, x_1).$
\end{definition}
\begin{property}[Interpolating point, \citep{ambrosio2005gradient}]
	Any point $x_t$ from a constant speed geodesic $(x(t))_{t\in[0,1]}$ is an {\em interpolating point} and verifies, $d(x_0,x_1)=d(x_0,x_t)+d(x_t,x_1),$ i.e. the triangle inequality becomes an equality.
\end{property}
These definitions and properties carry over to the case of the Wasserstein distance:
\begin{definition}[Wasserstein Geodesics, Interpolating measure, \citep{ambrosio2005gradient,kolouri2017optimal}]  
Let $\mu_0$, $\mu_1$ $\in \mathscr{P}_p(X)$ with 
$X \subseteq \R^d$ compact, convex and equipped
with $\mathcal{W}_p$. Let $\gamma \in \Pi(\mu_0, \mu_1)$ be 
an optimal transport plan.
For $t\in[0,1],$ let $\mu_t \doteq (\pi_t)_\#\gamma$ where $\pi_t(x, y) \doteq (1 - t)x + ty$,
i.e. $\mu_t$ is the push-forward measure of $\gamma$ under the map $\pi_t$.
Then, the curve $\bar\mu\doteq (\mu_t)_{t\in[0,1]}$ is a constant speed geodesic between $\mu_0$ and 
$\mu_1;$ we call it a {\em Wasserstein geodesics} between $\mu_0$ and 
$\mu_1.$

Any point $\mu_t$ of the geodesics is an {\em interpolating measure} between $\mu_0$ and $\mu_1$ and, as expected:
\begin{equation}
	\label{eq:interpolating_measure}
	\mathcal{W}_p(\mu_0,\mu_1)=\mathcal{W}_p(\mu_0,\mu_t)+\mathcal{W}_p(\mu_t,\mu_1).
\end{equation}
\end{definition}

In the discrete case, and for a fixed $t$, one can obtain such interpolating measure 
$\mu_t$ given the optimal transport map $\P^\star$ solution of
\Cref{eq:wd} as follows \cite[Remark 7.1]{peyre2019computational}:
\begin{equation}
    \label{eq:interpolant}
    \mu_t = \sum_{i,j}^{n,m} \P_{i,j}^\star \delta_{(1-t)x_i + t x_j^\prime}
\end{equation}
where $\P_{i,j}^\star$ is the $(i,j)$-th entry of $\P^\star$; as an interpolating measure, 
$\mu_t$ obviously complies with~\eqref{eq:interpolating_measure}.

\begin{figure}
    
 \begin{minipage}[t]{0.475\textwidth} 
            \centering
        \includegraphics[width=0.85\linewidth]{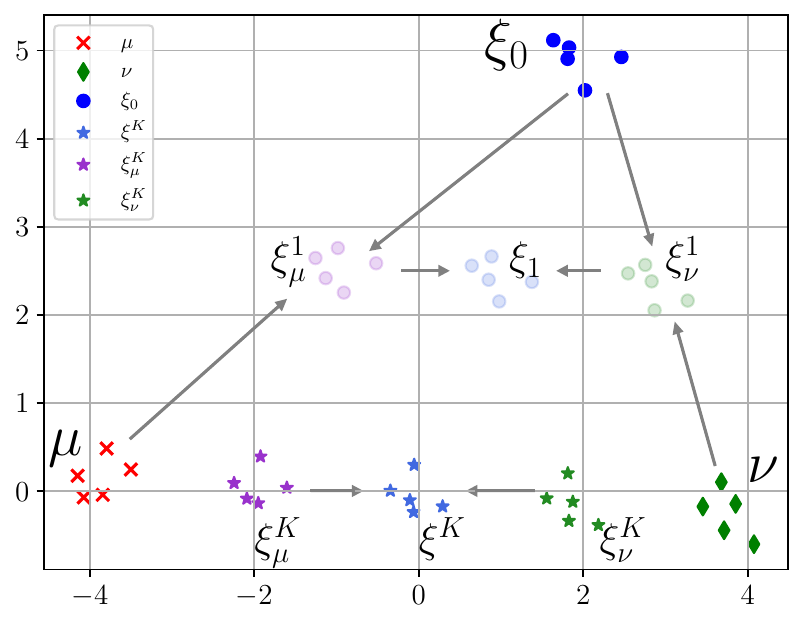}
                    \end{minipage}  
 \begin{minipage}[t]{0.475\textwidth} 
    \small
    \vspace{-4.2cm}
\begin{itemize}
    \item At first iteration the current estimation $\xi_0$ of $\xi^\star$ is sent to each client in order to
     compute two interpolating measures $\xi_\mu^1$ and $\xi_\nu^1$, which are sent back to the server.
    \item Server then computes an interpolating measure between $\xi_\mu^1$ and $\xi_\nu^1$ to obtain the 
    next iterate of the geodesic element  $\xi^1$
    \item The process is repeated until convergence to obtain $\xi^K$ and we define 
    $\mathcal{W}_p(\mu, \nu) = \mathcal{W}_p(\mu, \xi^K) + \mathcal{W}_p(\nu, \xi^K)$.
    
\end{itemize}

\end{minipage}
\caption{The Wasserstein distance between $\mu$ and $\nu$ which are on their respective clients
can be computed as $\mathcal{W}_p(\mu, \nu) = \mathcal{W}_p(\mu, \xi^\star) + \mathcal{W}_p(\nu, \xi^\star)$ where 
$\xi^\star$ is an element on the geodesic between $\mu$ and $\nu$. \ours seeks at
estimating  $\xi^{\star}$ with $\xi^K$ using an iterative algorithm and
plugs in this estimation to obtain $\mathcal{W}_p(\mu, \nu)$. 
Iterates of $\xi_i$ are computed on the server and sent to clients
in order to compute  measures $\xi_\mu^i$ and $\xi_\nu^i$ that are on geodesics of $\mu$ and $\xi_i$
and $\nu$ and $\xi_i$ respectively.}
\label{fig:fig0illustration}
\end{figure}

\subsection{Problem Statement}

Our goal is to compute the Wasserstein distance between two 
data distributions $\mu$ and $\nu$ on a global server with the constraint that $\mu$ and $\nu$ 
are distributed on two different clients which do not share any data samples to the server.
From a mathematical point of view, our objective is to estimate an element $\xi^\star$ on the geodesic
of $\mu$ and $\nu$ without having access to them by leveraging two other
elements $\xi_\mu$ and $\xi_\nu$ on the geodesics of $\mu$ and $\xi^\star$ and $\nu$ and $\xi^\star$ respectively. 

\subsection{Related Works} 
Our work touches the specific question of learning/approximating a distance between distributions
whose samples are secluded on isolated clients. As far as we are aware of, this is a problem that 
has never been investigated before and there are only few works that we see closely
connected to ours.
Some problems have addressed the objective of retrieving
nearest neighbours of one vector in a federated manner. For instance, \cite{liu2021secure} consider to exchange encrypted versions of the dataset
on client to the central server and
 \cite{Schoppmann2018PrivateNN} 
 consider the exchange of differentially private statistics about the client dataset.
 \cite{zhang2023approximate} propose a federated approximate
 $k$-nearest approach based on a specific spatial data federation.
 Compared to these works that compute distances in a federated manner, 
 we address the case of distances on distributions without any specific
 encryption of the data and we exploit the properties of the 
 Wasserstein distances and its geodesics, which have been overlooked in the mentioned works.
If these properties have been relied upon as a key tool in some computer vision applications \citep{bauer2015diffeomorphic,maas2017transport}
 and trajectory inference \citep{huguet2022manifold}, they have
 not been employed as a privacy-preserving tool.

\section{Computing the Federated Wasserstein distance}
\label{sec:contribution}

In this section, we develop a methodology to compute, on a global server, the Wasserstein distance between two 
distributions $\mu$ and $\nu$, stored on
two different clients which do not share this information to the server. Our approach leverages the topology induced by the Wasserstein distance in the space of probability measures, and more precisely, the geodesics.

\textbf{Outline of our methodology.} A key property is that $\mathcal{W}_p$ is a metric, thus satisfies the triangle inequality: for any $\mu, \nu, \xi \in \mathscr{P}_p(X)$,
 \begin{align}
    \label{eq:triangle}
    \mathcal{W}_p(\mu, \nu) \leq \mathcal{W}_p(\mu, \xi) + \mathcal{W}_p(\xi, \nu) \,,
 \end{align}
 with equality if and only if $\xi = \xi^\star$, where $\xi^\star$ is an interpolating measure. 
 Consequently, one can compute $\mathcal{W}_p(\mu, \nu)$ by computing $\mathcal{W}_p(\mu, \xi^\star)$ and 
 $\mathcal{W}_p(\xi^\star, \nu)$ and adding these two terms. This result is 
  useful in the federated setting and inspires our methodology, as described hereafter. The global server computes $\xi^\star$ and communicate $\xi^\star$ to the two clients. The clients respectively compute $\mathcal{W}_p(\mu, \xi^\star)$ and $\mathcal{W}_p(\xi^\star, \nu)$, then send these to the global server. Finally, the global server adds the two received terms to return $\mathcal{W}_p(\mu, \nu)$.

 The main bottleneck of this procedure is that the global server needs to compute $\xi^\star$ (which by definition, depends on $\mu, \nu$) while not having access to $\mu, \nu$ (which are stored on two clients). We then propose a simple workaround to overcome this challenge, based on an additional application of the triangle inequality: for any $\xi \in \mathscr{P}_p(X)$,
 \begin{equation}
       \mathcal{W}_p(\mu, \nu) \leq \mathcal{W}_p(\mu, \xi) + \mathcal{W}_p(\xi, \nu) = \mathcal{W}_p(\mu, \xi_\mu) + 
   \mathcal{W}_p(\xi_\mu, \xi)+ \mathcal{W}_p(\xi, \xi_\nu) + \mathcal{W}_p(\xi_\nu, \nu) \,, \label{eq:interp0}
 \end{equation}
 where $\xi_\mu$ and $\xi_\nu$ are interpolating measures respectively between 
 $\mu$ and $\xi$ and $\xi$ and $\nu$. Hence, computing $\xi^\star$ can be done through intermediate measures $\xi_\mu$ and $\xi_\nu$, to ensure that $\mu, \nu$ stay on their respective clients. To this end, we develop an optimization procedure which essentially consists in iteratively estimating an interpolating measure $\xi^{(k)}$ between $\mu$ and $\nu$ on the server, by using $\xi_{\mu}^{(k)}$ and $\xi_{\nu}^{(k)}$ which were computed and communicated by the clients. More precisely, the objective is to minimize \eqref{eq:interp0} over $\xi$ as follows: at iteration $k$, the clients receive current iterate $\xi^{(k-1)}$ and compute $\xi_{\mu}^{(k)}$ and $\xi_{\nu}^{(k)}$ (as interpolating measures between $\mu$ and $\xi^{(k-1)}$, and between $\xi^{(k-1)}$ and $\nu$ respectively). By the triangle inequality,
    \begin{equation}
    \mathcal{W}_p(\mu, \nu) \leq \mathcal{W}_p(\mu, \xi_{\mu}^{(k)}) + \mathcal{W}_p(\xi_{\mu}^{(k)}, 
    \xi^{(k-1)})+ \mathcal{W}_p(\xi^{(k-1)}, \xi_{\nu}^{(k)}) + \mathcal{W}_p(\xi_{\nu}^{(k)}, \nu) \,, \label{eq:interp}
\end{equation}
therefore, the clients then send $\xi_{\mu}^{(k)}$ 
and $\xi_{\nu}^{(k)}$ to the server, which in turn, computes the next iterate $\xi^{(k)}$
by minimizing the left-hand side term of \eqref{eq:interp}, \ie
\begin{equation}
   \xi^{(k)} \in \argmin_\xi   \mathcal{W}_p(\xi_{\mu}^{(k)}, 
   \xi)+ \mathcal{W}_p(\xi, \xi_{\nu}^{(k)}) \,. \label{eq:mininterp}       
\end{equation}

\begin{wrapfigure}{R}{0.5\textwidth}
    \vspace{-0.4cm}
    \begin{minipage}{0.5\textwidth}
\begin{algorithm}[H]
    \footnotesize
	\caption{\textcolor{magenta}{\texttt{\ours}}}
	\label{algo:fedOT}
	\begin{algorithmic}[1]
		\REQUIRE{$\mu$ and $\nu$, initialisation of $\xi^{(0)}$,
        function \emph{InterpMeas} that computes an interpolating measure between two measures using  \Cref*{eq:interpolant} or 
        \Cref*{eq:approxinterpolant} for any $0<t<1$.}
		\FOR{$k=1$ {\bfseries to} $K$}
            \STATE {\textcolor{darkgreen}{\textit{// Send $\xi^{(k-1)}$ to clients}}}
            \STATE {\textcolor{darkgreen}{\textit{// Compute on clients with optional return of the distances}}}
                        			\STATE $\xi_{\mu}^{(k)}, [d_{\mu,\xi^{(k)}}] \gets$ {\texttt{InterpMeas}}$(\mu, \xi^{(k-1)})$  
            \STATE $\xi_{\nu}^{(k)}, [d_{\xi^{(k),\nu}}] \gets$ {\texttt{InterpMeas}}$(\nu, \xi^{(k-1)})$ 
            \STATE {\textcolor{darkgreen}{\textit{// Send $\xi_{\mu}^{(k)}$  and $\xi_{\nu}^{(k)}$ to server}}}
            \STATE $\xi^{(k)} \gets$  {\texttt{InterpMeas}}$(\xi_{\mu}^{(k)}, \xi_{\nu}^{(k)})$
		\ENDFOR
        \STATE {\textcolor{darkgreen}{\textit{// Send $d_{\mu,\xi^{(k)}} + d_{\xi^{(k),\nu}}$ to server}}}
        \STATE  $d_{\mu,\nu} = d_{\mu,\xi^{(K)}} + d_{\xi^{(K)},\nu}$

		\ENSURE return  $d_{\mu,\nu}$ on server
\end{algorithmic}
\end{algorithm}
\end{minipage}
\end{wrapfigure}
Our methodology is illustrated in \Cref*{fig:fig0illustration} and summarized in \Cref{algo:fedOT}. Besides computing the Wasserstein distance in a federated manner, we point out several methods
 can easily be incorporated in our algorithm to further reduce the risk of privacy leak.
  Since the triangle inequality reaches equality on a particular geodesic, 
  $\xi^{(k)}$, $\xi_{\mu}^{(k)}$ or $\xi_{\nu}^{(k)}$ are not unique,
 thus clients can compute these interpolating measures based on a \emph{random} value of $t$. Besides, 
 since communicating the distance may reveal information about the data, they can be
 shared with the server only for the last iteration.   More effectively, one can incorporate an (adapted) differentially private version of the Wasserstein 
  distance \citep{le2019differentially}.
  Regarding communication cost, at each iteration, the communication cost involves the transfer
  between the server and the clients of four interpolating measures: 
  $\xi^{(k-1)}$ (twice), $\xi_\mu^{(k)}$, $\xi_\nu^{(k)}$. Hence, 
  if the support size of $\xi^{(k-1)}$ is $S$, the communication cost is in $\mathcal{O}(4SKd)$,
  with $d$ the data dimension and $K$ the number of iterations.

\paragraph{Reducing the computational complexity.}
In terms of computational complexity, we need to compute three OT plans per iteration which 
single cost, based on the network simplex is $O((n+m)nm log(n+m))$.
More importantly, consider that $\mu$ and $\nu$ are discrete measures, 
then, any interpolating measure between $\mu$ and $\nu$ is supported on at most on $n+m+1$ points.
Hence, even if the size of the support of $\xi^{(0)}$ is small, but $n$ is large, 
the support of the next interpolating measures may get larger and larger,
and this can yield an important computational overhead
 when computing $\mathcal{W}_p(\mu, \xi^{(k)})$ and $\mathcal{W}_p(\xi^{(k)}, \nu)$. 

 To reduce this complexity, we resort to approximations of the interpolating measures
 which goal is to fix the support size of the interpolating measures to a small number $S$.
 The solution we consider is to approximate the McCann's interpolation equation 
which formalizes
geodesics $\xi_t$ given an optimal transport map between two distributions,say, $\xi$ and $\xi^\prime$, 
based on the equation 
$\xi_t = ((1-t)Id + tT)_\# \xi$ \cite{peyre2019computational}. 
Using barycentric mapping approximation 
of the map $T$ \citep{courty2018learning},
we propose to approximate the interpolating measures $\xi_t$ as 
\begin{equation}
\label{eq:approxinterpolant}
\xi_t = \frac{1}{n}\sum_{i=1}^n \delta_{(1-t)x_i + t n(\P^\star \mathbf{X}^\prime)_i}  
\end{equation}
where $\P^\star$ is the optimal transportation plan between $\xi$ and $\xi^\prime$, $x_i$ and
 $x_j^\prime$ are the samples from these distributions and $\mathbf{X}^\prime$ is the matrix
    of samples from $\xi^\prime$.
 Note that by choosing
 the appropriate formulation of the equation, the  support size of 
 this interpolating measure can be chosen as the one of $\xi$ or $\xi^\prime$. In practice,
 we always opt for the choice that leads to the smallest support of 
 the interpolating measure. Hence, if the support size of  $\xi^{(0)}$ is $S$, we have the guarantee
that the support of $\xi^{(k)}$ is $S$ for all $k$. 
Then, for computing $\mathcal{W}_p(\mu, \xi^{(k)})$ using approximated interpolating measures,
 it costs $O(3*(Sn^2 + S^2n)log(n+S))$ at each iteration and if $S$ and the number of 
   iterations $K$ are small enough, the approach we propose is even competitive compared to exact OT. 
   Our experiments reported later that for larger number 
   of samples ($\geq 5000$), our approach is as fast as exact optimal transport and
    less prone to numerical errors.

\paragraph{Theoretical guarantees.} 
We discuss in this section some theoretical properties  of the components of \ours. At 
first, we show that the approximated interpolating measure is tight in the sense that there
exists some situations where the resulting approximation is exact.

\begin{theorem} 
    Consider two  discrete distributions  $\mu$ and $\nu$ with the same number of samples
     $n$ and uniform weights,
     then for any $t$, the approximated interpolating measure, between $\mu$ and $\nu$ 
     given by 
    \Cref{eq:approxinterpolant} is equal to the exact one \Cref{eq:interpolant}.
    \end{theorem}

Proof is given in \Cref*{sec:proof_approx}. In practice, this property does not have much impact, but it ensures us about the soundness
of the approach. In the next theorem, we prove that \Cref{algo:fedOT} is 
theoretically justified, in the sense that its output converges to $\mathcal{W}_p(\mu,\nu)$.

\begin{theorem} \label{thm:convergence}
    Let $\mu$ and $\nu$ be two measures in $\mathscr{P}_p(X)$, 
    $\xi_{\mu}^{(k)}$, $\xi_{\nu}^{(k)}$
    and $\xi^{(k)}$ be the interpolating measures computed at iteration $k$ as defined
    in \Cref{algo:fedOT}.
    Denote as 
    $$A^{(k)} = \mathcal{W}_p(\mu, \xi_{\mu}^{(k)}) + \mathcal{W}_p(\xi_{\mu}^{(k)}, 
    \xi^{(k)})+ \mathcal{W}_p(\xi^{(k)}, \xi_{\nu}^{(k)}) + \mathcal{W}_p(\xi_{\nu}^{(k)}, \nu)
    $$ 
    Then the sequence $(A^{(k)})_k$ is non-increasing and converges to $\mathcal{W}_p(\mu, \nu)$.
\end{theorem}

We provide hereafter a sketch of the proof, and refer to \Cref{sec:proof1} for full details. First, we show that the sequence $(A^{(k)})_k$ is non-increasing, as we iteratively 
update $\xi_\mu^{(k+1)}$, $\xi_\nu^{(k+1)}$ and $\xi^{(k+1)}$ based on geodesics (a minimizer
of the triangle inequality). Then, we show that the sequence $(A^{(k)})_k$ is bounded below by $\mathcal{W}_p(\mu, \nu)$. We conclude the proof by proving that the sequence $(A^{(k)})_k$ converges to $\mathcal{W}_p(\mu, \nu)$.

In the next theorem, we show that when $\mu$ and
$\nu$ are Gaussians then we can recover some nicer properties of our algorithm
and provide a convergence rate (proof in \Cref*{sec:proof_convergence}).

\begin{theorem}
    Assume that $\mu$, $\nu$ and $\xi^{(0)}$ are three Gaussian distributions with the same covariance matrix $\Sigma$
    ie $\mu \sim \mathcal{N}(\m_\mu, \Sigma)$, $\nu \sim \mathcal{N}(\m_\nu, \Sigma)$
    and $\xi^{(0)} \sim \mathcal{N}(\m_{\xi^{(0)}}, \Sigma)$. Further assume that we are not
    in the trivial case where $\m_\mu$, $\m_\nu$, and $\m_{\xi^{(0)}}$ are aligned.
    Applying our \Cref{algo:fedOT} with $t=0.5$ and the squared Euclidean cost, we have the following properties:
    \begin{enumerate}
      \item all interpolating measures $\xi_{\mu}^{(k)}$,$\xi_{\nu}^{(k)}$, $\xi^{(k)}$ are Gaussian distributions with
       the same covariance matrix $\Sigma$,
      \item for any $k \geq 1$, $\mathcal{W}_2(\mu,\nu) = \| \m_\mu - \m_\nu\|_2 = 2 \| \m_{\xi_\mu^{(k)}} - \m_{\xi_\nu^{(k)}}\|_2 
      = 2 \mathcal{W}_2(\xi_\mu^{(k)},\xi_\nu^{(k)}) $
      \item $\mathcal{W}_2(\xi^{(k)},\xi^{\star}) = \frac{1}{2} \mathcal{W}_2(\xi^{(k-1)},\xi^{\star})$ 
      \item $\mathcal{W}_2(\mu,\xi^{(k)}) + \mathcal{W}_2(\xi^{(k)},\nu) - \mathcal{W}_2(\mu,\nu)\leq \frac{1}{2^{k-1}} \mathcal{W}_2(\xi^{(0)},\xi^{(\star)})$
    \end{enumerate}
  \end{theorem}
  Interestingly, this theorem also says that in this specific case, only one iteration is needed to recover 
  $\mathcal{W}_2(\mu,\nu)$

\begin{figure}[t]
    ~\hfill~\includegraphics[width=3.5cm]{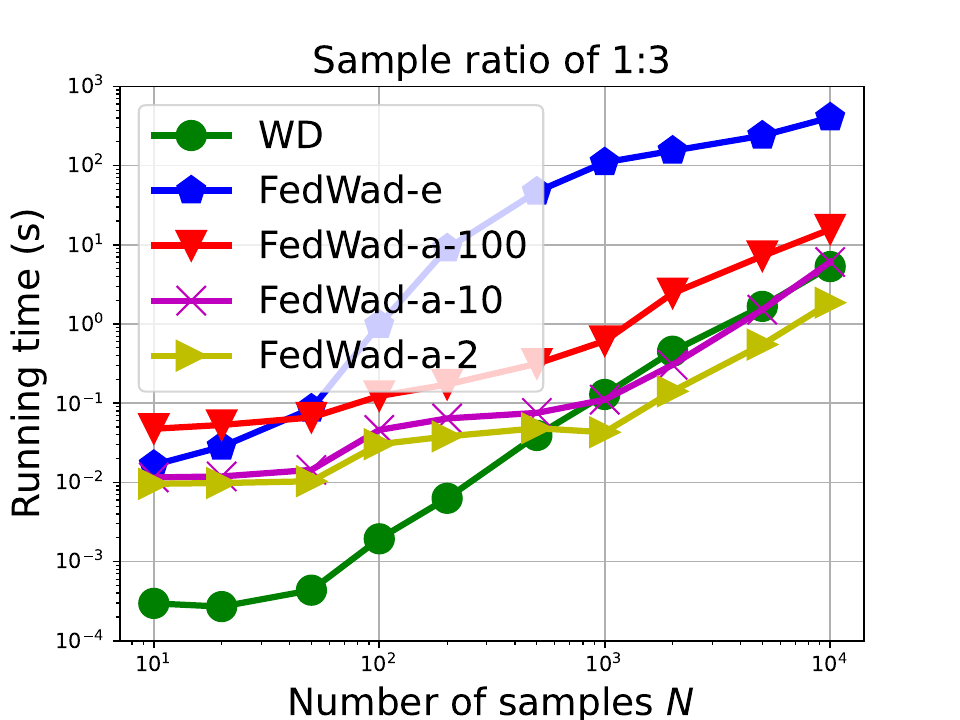}~\hfill~
    \includegraphics[width=3.5cm]{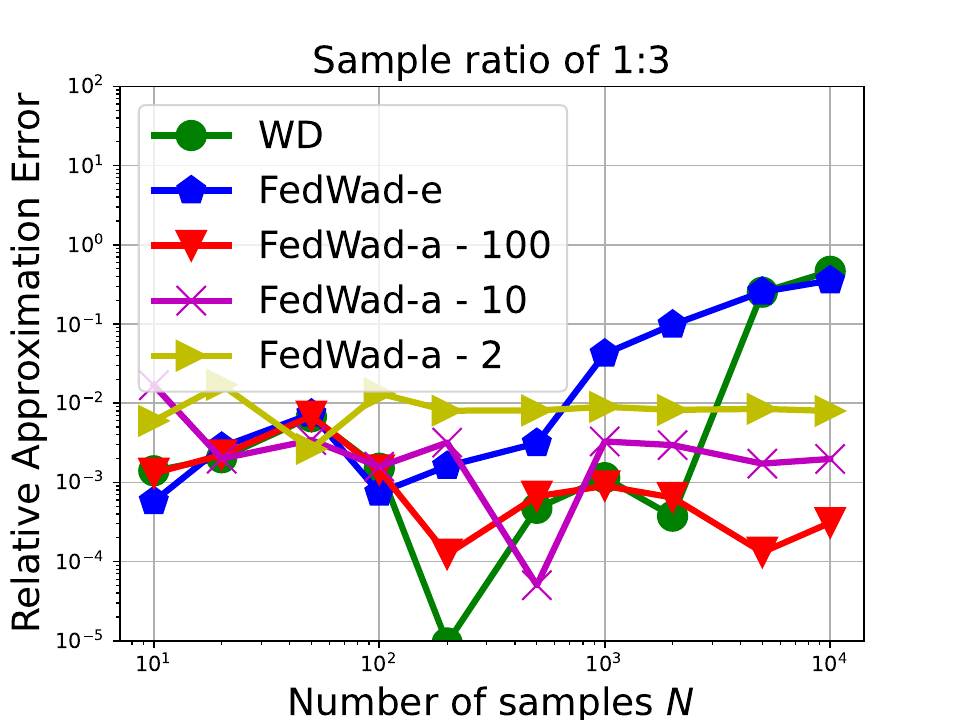}~\hfill~
    \includegraphics[width=3.5cm]{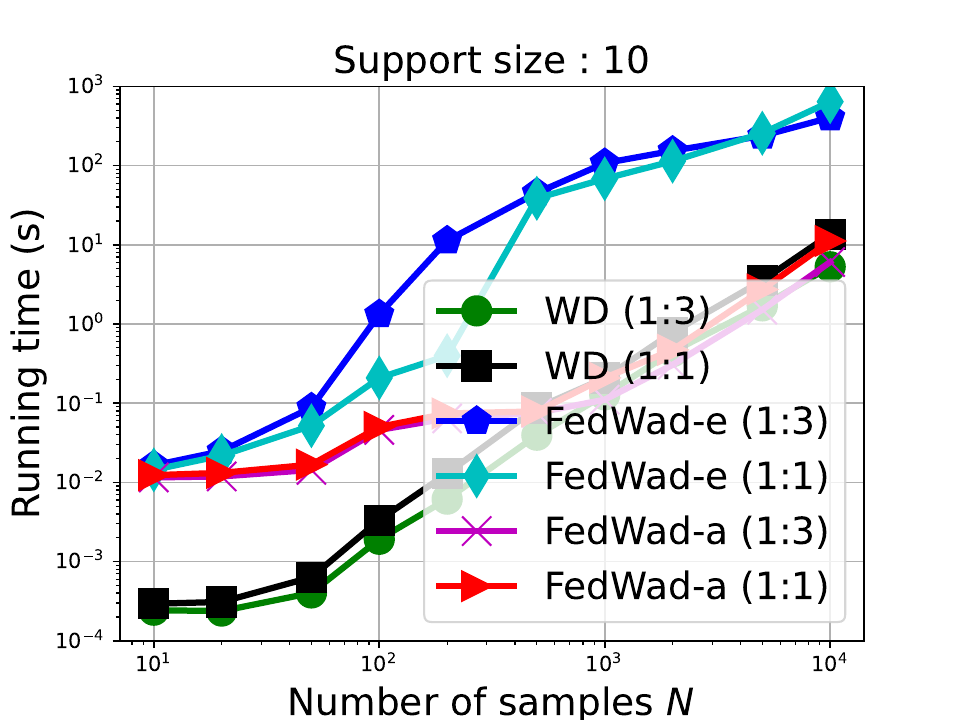}~\hfill~
    \includegraphics[width=3.5cm]{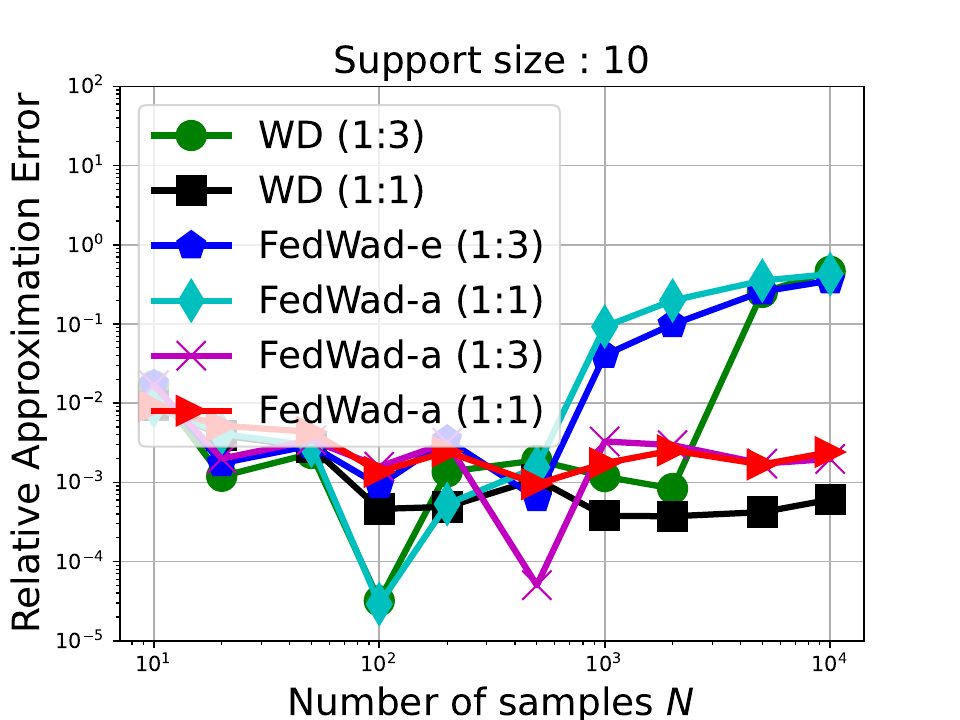}~\hfill~

    \caption{Analysis of the different Wasserstein distance computation methods (most-left panels) 
    for varying support size of the approximated \ours
    and (most-right panels) for varying sample ratio in the two distributions
    and fixed support size.
    For each couple of panels, for increasing number of samples, 
    we report the running time and the relative error of the Wasserstein distance (WD), our exact FedWaD (FedWad-e) and
     our approximate FedWaD (FedWad-a) with a support
     size of $2$, $10$ and $100$. 
     For the most-right panels, we have set the support size of the interpolating measure to $10$.
     For a sample ratio (1:3), the first 
     distribution has a number of samples $N$ and the second ones $N/3$. }
    \label{fig:approximate}
  \end{figure}
  \newpage
  \section{Experiments}
  \label{sec:results}

This section presents  numerical applications, where \ours can successfully be used
and show how it can boost performances of federated learning algorithms. Full details
are provided in \Cref*{app:expe}.

\begin{wrapfigure}{r}{0.5\textwidth}    \centering
    \includegraphics[width=0.235\textwidth]{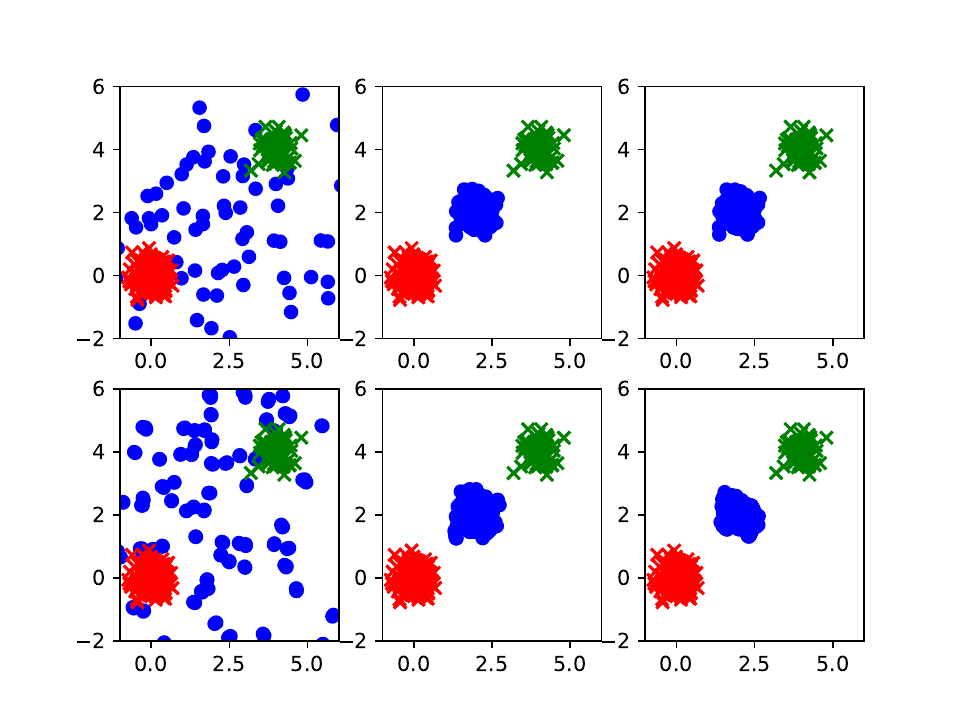}
    \includegraphics[width=0.235\textwidth]{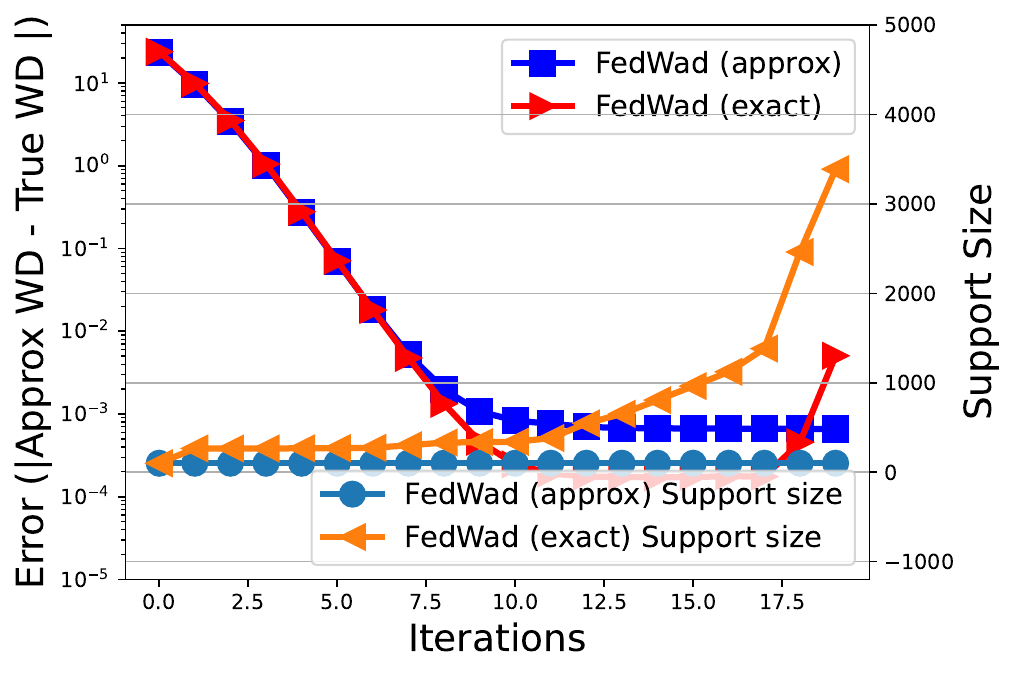}
        \caption{ (left) Evolution of the interpolating measure $\xi^{(k)}$ - in blue -  (right) the estimated Wasserstein distance 
    between two Gaussian distributions $\mu$ and $\nu$. 
                    }
    \label{fig:toy}
\end{wrapfigure}

\paragraph{Toy analysis.}
We illustrate the evolution of interpolating measures using \ours\
for calculating the Wasserstein distance between two Gaussian distributions. 
We sample 200 points from two 2D Gaussian distributions with different means and the same covariance matrix. 
We compute the interpolating measure at  $t=0.5$ using both the analytical formula \eqref{eq:interpolant} and the approximation
\eqref{eq:approxinterpolant}. \Cref{fig:toy} (left panel) shows how the interpolating measure evolves across iterations.
We also observe,  in \Cref{fig:toy} (right panel), that the error on the true Wasserstein distance for
the approximated interpolating measure reaches $10^{-3}$, while for the exact interpolating measure,
 it drops to a minimum of $10^{-4}$  before increasing.
This discrepancy occurs as the support size of the interpolating measure expands across iterations
 leading to numerical errors when computing the optimal transport plan between
$\xi^{(k)}$ and $\xi_\mu^{(k)}$ or $\xi_\nu^{(k)}$.
Hence, using the 
approximation \Cref{eq:approxinterpolant} is a more robust alternative to exact 
computation \Cref{eq:interpolant}.

We also examine computational complexity and approximation errors for both methods as
 we increase sample sizes
 in the distributions, as displayed in \Cref{fig:approximate}. Key findings include:
The approximated interpolating measure significantly improves computational efficiency,
 being at least 10 times faster with sample size exceeding 100, especially with smaller support sizes.
It also achieves a similar relative 
approximation error as \ours using the exact interpolating measure and true non-federated Wasserstein distance.
 Importantly, it demonstrates greater robustness with larger 
sample sizes compared to  true Wasserstein distance for such a small dimensional problem.

\paragraph{Wasserstein coreset and application to federated learning.} In many ML applications,
  \begin{wrapfigure}{r}{0.4\textwidth} \vspace{-0.5cm}   
    \begin{center}
    \includegraphics[width=0.3\textwidth]{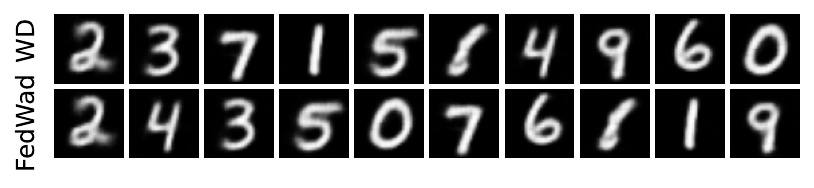}
    \includegraphics[width=0.3\textwidth]{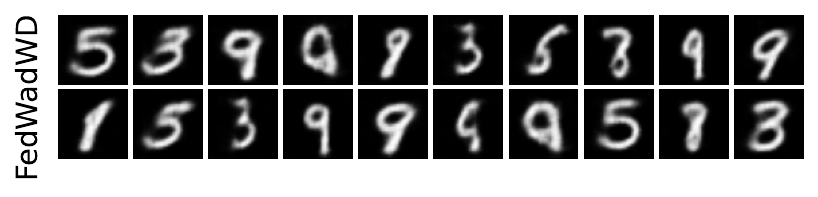}
    \includegraphics[width=0.3\textwidth]{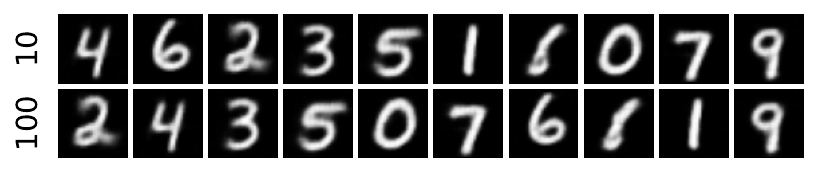}~\\

    \end{center}
     \caption{Examples of the $10$ coreset we obtained, with for each panel \emph{(top-row)} the exact Wasserstein
     and \emph{(bottow-row)} \ours for the \texttt{MNIST} dataset. Different panels correspond to different number of classes $K$ on each client: \emph{(top)} 
     $K=8$, \emph{(middle)} $K=2$, \emph{(bottom)} support of the interpolating measure
     varying from $10$ to $100$.}
     \label{fig:coreset}
\end{wrapfigure}
 summarizing data into fewer representative samples is routinely done to deal with large datasets. 
The notion of \emph{coreset} has been relevant to extract such samples and admit several formulations
\citep{phillips2016coresets,munteanu2018coresets}. In this experiment, we show that Wasserstein coresets 
\citep{claici2018wasserstein} can be computed in a federated way via \ours. Formally, 
given a dataset described by the  distribution $\mu$, 
the Wasserstein coreset aims at finding the empirical distribution
that minimizes
$\min_{x_1^\prime, \cdots,x_K^\prime } \mathcal{W}_p\left(\frac{1}{K}\sum_{i=1}^K \delta_{x_i^\prime}, \mu\right) \,.
$
We solve this problem in the following federated setting: we assume
 that either the samples drawn from $\mu$ are stored on an unique client 
 or distributed across different clients, and the objective is to learn the coreset samples
  $\{x_i^\prime\}$ on the server. 
  In our setting, we can compute the federated Wasserstein distances between
the current coreset and some subsamples of all active client datasets, then update the coreset
  given the aggregated gradients of these distances with respect to the coreset support. 
We sampled $20000$ examples randomly from the \texttt{MNIST} dataset,
and dispatched them at random on $100$ clients.
We compare the results we obtained with \ours with those obtained with exact non-federated Wasserstein distance 
The results are shown in \Cref{fig:coreset}. We can note that when classes are 
almost equally spread across clients (with $K=8$ different classes per client),
 \ours is able to capture the $10$ modes of the dataset. However, as the diversity in classes
 between clients increases, \ours has more difficulty to capture all the modes of 
 the dataset. Nonetheless, we also observe that the exact Wasserstein distance is not
 able to recover those modes either. We can thus conjecture that this failure is likely
 due to the coreset approach itself, rather than to the approximated distance returned by \ours. 
 We also note that the support size of the interpolating measure has less impact
 on the coreset. We believe this is a very interesting result, as it shows that \ours can
 provide useful gradient to the problem even with a poorer estimation of the distance. 

\begin{wrapfigure}{r}{0.4\textwidth}\vspace{0.75cm}   
  \begin{center}
  \includegraphics[width=4.5cm]{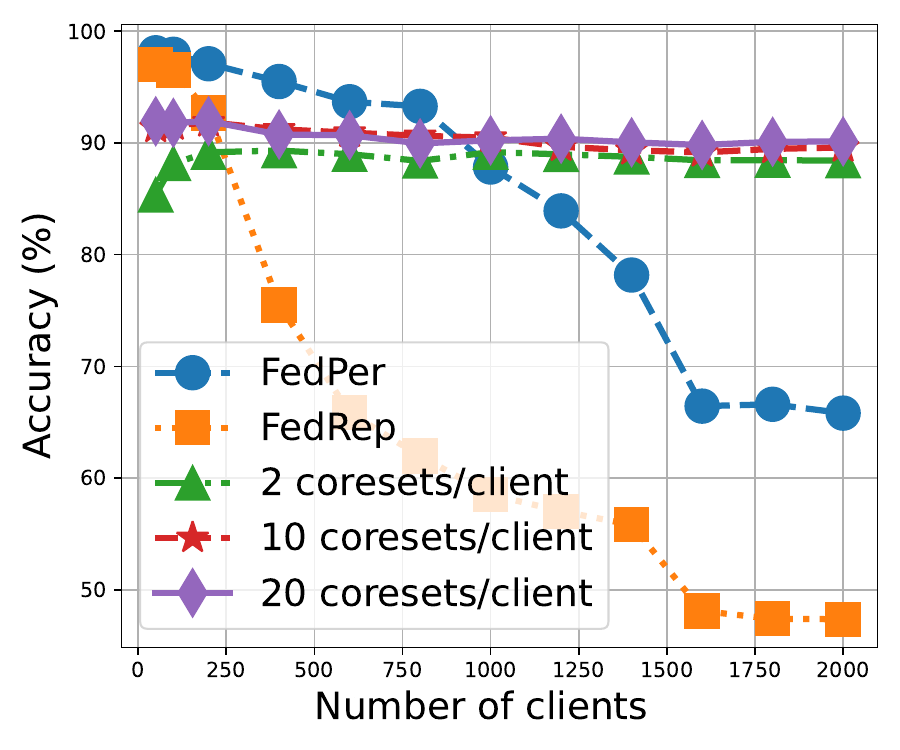} 
  \end{center}
  \vspace{-0.5cm}
  \caption{ Nearest neighbor classifier based on the coresets learnt from each client for varying number of clients and 
  number of coresets per clients We have compared to the performance of two personalized FL algorithms.}
  \label{fig:coreset_perf}
\end{wrapfigure}

\paragraph{Federated coreset classification model} Those federated coresets can also be used for classification tasks. As such, we have learned
coresets for each client, and used all the coresets from all clients as the examples for a 
one-nearest neighbor global classifier shared to all clients. Note that since a coreset computation is an unsupervised task,
we have assigned to each element of a coreset the label of the closest element in 
the client dataset. 
For this task, we have used the \texttt{MNIST} dataset which has been autoencoded in order
to reduce its dimensionality. Half of the training samples have been used for learning the autoencoder
and the other half for the classification task. Those samples and the test samples of dataset
have been distributed across  clients while ensuring that each client has samples from only $2$ 
classes.
We have then computed the accuracy of this federated classifier for varying number of clients
and number of coresets and compared its performance to the ones of \emph{FedRep} \citep{pmlr-v139-collins21a} and 
\emph{FedPer} \citep{arivazhagan2019federated}. Results are reported in \Cref{fig:coreset_perf}. We can see that
our simple approach is highly competitive with these personalized FL approaches, and even outperforms them
when the number of users becomes large. 

\paragraph{Geometric dataset distances via federated Wasserstein distance.}

Our goal is to improve on the seminal algorithm of \cite{alvarez2020geometric}
that seeks at computing distance between two datasets $\mathcal{D}$ and $\mathcal{D}^\prime$ using optimal transport.
We want to make it federated.
 This extension will pave the way to better federated learning algorithms for transfer learning and domain adaptation or can
simply be used for boosting federated learning algorithms, as we illustrate next. 
\cite{alvarez2020geometric} 
considers a Wasserstein distance with a ground metric that mixes
distances between features and tractable distance between class-conditional distributions.
For our extension, we will use the same ground metric, but we will compute the Wasserstein
 distance using \ours. Details are provided in \Cref*{app:otdd}.

\begin{figure}
    \begin{center}
    \includegraphics[width=0.98\textwidth]{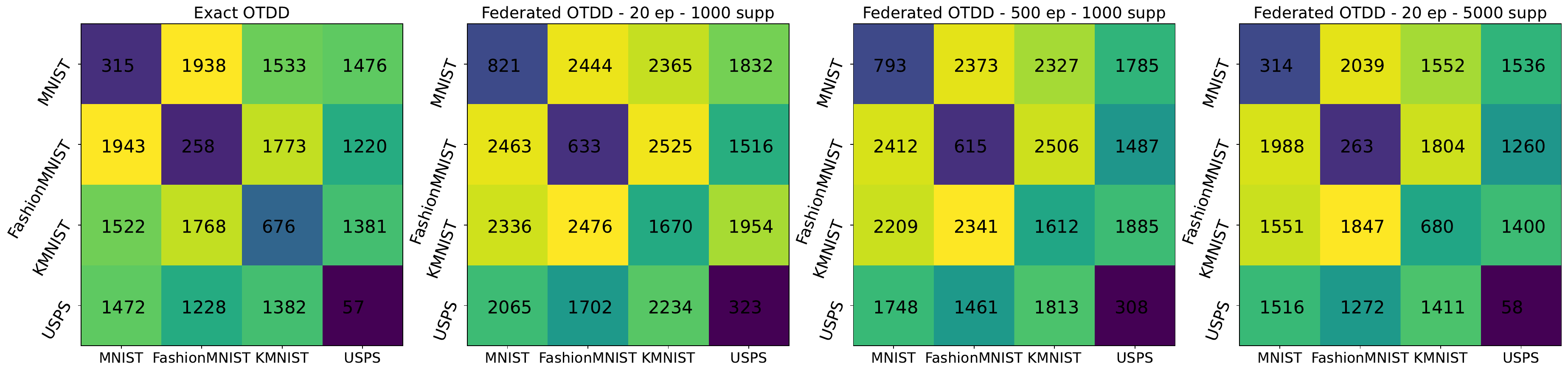}
    \end{center}
    \caption{Comparison of the matrix of distances between digits datasets computed by \ours and 
    the true OTDD distance between the same datasets. \emph{(left)} OTDD, \emph{(middle-left)} \ours with $20$ epochs
    and $1000$ support points,
    \emph{(middle-right)} \ours with $500$ epochs and $1000$ support points, \emph{(right)} \ours with $20$ epochs 
    and $5000$ support points}
    \label{fig:otddapproxdistancemat}
\end{figure}

We replicated the experiments of \cite{alvarez2020geometric} on the
dataset selection for transfer learning: given a source dataset,
the goal is to find a target one which is the most similar to the source. 
 We considered four real datasets, namely \texttt{MNIST}, \texttt{KMNIST},
 \texttt{USPS} and \texttt{FashionMNIST} and we have computed all the pairwise
 distance between $5000$ randomly selected examples from each dataset using the
 original \texttt{OTDD} of \cite{alvarez2020geometric} and our \ours approach. 
For \ours, we chose the support size of the interpolating measure to be
$1000$ and $5000$ and the number of epochs to $20$ and $500$. Results, averaged over
$5$ random draw of the samples, are
depicted in \Cref{fig:otddapproxdistancemat}. We can see that the distance matrices
produced by \ours are semantically similar to the ones for OTDD distance, which means
that order relations are well-preserved for most pairwise distances (except only for two
pairs of datasets in the \texttt{USPS} row). More importantly, running more epochs
leads to slightly better approximation of the OTDD distance, but the exact order relations are already
uncovered using only $20$ epochs in \ours.
Detailed ablation studies on these parameters are provided in \Cref*{app:fedotddanalysis}.

\paragraph{Boosting FL methods} One of the challenges in FL is the heterogeneity of the data distribution among clients. 
This heterogeneity is usually due to shift in class-conditional distributions or to a label 
shift (some classes being absent on a client).
As such, we propose to investigate a simple approach that allows to address  dataset heterogeneity (in terms of distributions)
among clients, by leveraging on our ability to compute distance between datasets in a federated way.

Our proposal involves computing pairwise dataset distances between clients,
clustering them based on their (di)-similarities using a spectral clustering 
algorithm \citep{von2007tutorial}, and using this clustering knowledge 
to enhance existing federated learning algorithms. 
In our approach, we run the FL algorithm for each of the $K$ clusters of clients
instead of all clients to avoid information exchange between clients with  diverse datasets.
For example, for FedAvg, this means learning a global model for each cluster of clients, 
resulting in $K$ global models. For personalized models like FedRep \citep{pmlr-v139-collins21a}, 
or FedPer \citep{arivazhagan2019federated}, 
we  run the personalized algorithm on each cluster of clients.  By running FL algorithms
on clustered client, we ensure  
information exchange only between similar clients and improves the overall performance of
federated learning 
algorithms {by reducing the statistical dataset heterogeneity among clients}.

We have run experiments on MNIST and CIFAR10 in which client datasets hold a clear  cluster
structure. 
 We have also run  experiments where there is no cluster structure in which clients are randomly assigned a pair of classes.
In practice, we used the code of FedRep \cite{pmlr-v139-collins21a} for the
\emph{FedAvg}, \emph{FedRep} and \emph{FedPer} and the spectral clustering
method of scikit-learn \citep{scikit-learn} (details are in \Cref*{sec:boosting_apdx}). 
Results are reported in \Cref{tab:clustering} (with details in \Cref*{sec:boosting_apdx}). We can see that when there is a clear clustering structure
among the clients, \ours  is able to recover it and always improve the performance of the
original federated learning algorithms. Depending on the algorithm, the improvement can be highly significant.
For instance, for \emph{FedRep}, the performance can be improved by $9$ points for CIFAR10 and
up to $29$ for MNIST. ~ Interestingly, even without clear clustering structure, \ours is able to almost always improve the performance
of all federated learning algorithms (except for some specific cases of \emph{FedPer}).
Again for \emph{FedRep}, the performance uplift can reach $19$ points for CIFAR10 and $36$ for MNIST.
In terms of clustering, the  ``affinity" parameter of the spectral clustering algorithm seems to be the 
most efficient and robust one.

    \begin{table*}
        \caption{{MNIST}/CIFAR10 Average performances over 5 trials of three FL algorithms, FedAvg, FedRep and
        FedPer. For each algorithm we compare the vanilla performance with the ones
        obtained after clustering the clients using the \ours OTDD distance and three different
        setting of the spectral clustering algorithm (details in Appendix) and for a support size of $10$.
        The number of clients varies from $20$ to $100$. Bolded number indicate the
         best performing approach (and clustering parameters).  
        \label{tab:clustering}}
        \centering
        \tiny
        \ra{0.9}
        \begin{tabular}{@{}rrrrrrrrrr@{}}\toprule
         \multicolumn{1}{c}{~} & \multicolumn{4}{c}{Strong structure} & \multicolumn{1}{c}{~} & \multicolumn{4}{c}{No structure} \\  
        \cmidrule{2-5} \cmidrule{7-10}
        & \multicolumn{1}{c}{~} & \multicolumn{3}{c}{Clustering} && \multicolumn{1}{c}{~} & \multicolumn{3}{c}{Clustering} \\
         \cmidrule{3-5} \cmidrule{8-10}
        &  \multicolumn{1}{c}{Vanilla} & \multicolumn{1}{c}{Affinity}  & \multicolumn{1}{c}{Sparse G. (3)} &\multicolumn{1}{c}{Sparse G. (5)}  
        & & \multicolumn{1}{c}{Vanilla} & \multicolumn{1}{c}{Affinity}  & \multicolumn{1}{c}{Sparse G. (3)} &\multicolumn{1}{c}{Sparse G. (5)} \\
                        \midrule
        \rowcolor{Gray} \multicolumn{10}{c}{MNIST} \\\cmidrule{3-9}
        FedAvg\\
                20& 26.3 $\pm$ 3.8 & \textbf{99.5} $\pm$ \textbf{0.0} & \textbf{99.5} $\pm$ \textbf{0.0} & 91.5 $\pm$ 10.3 && 25.1 $\pm$ 6.6 & \textbf{71.3} $\pm$ \textbf{7.3} & 59.5 $\pm$ 3.0 & 57.0 $\pm$ 4.4 \\
        40& 39.1 $\pm$ 9.0 & \textbf{99.2} $\pm$ \textbf{0.1} & 91.1 $\pm$ 6.5 & 94.5 $\pm$ 9.4 && 42.5 $\pm$ 10.5 & \textbf{70.8} $\pm$ \textbf{13.5} & 60.0 $\pm$ 3.7 & 58.1 $\pm$ 6.3 \\
        100& 39.2 $\pm$ 7.7 & \textbf{98.9} $\pm$ \textbf{0.0} & 95.9 $\pm$ 4.6 & 98.4 $\pm$ 0.8 && 52.6 $\pm$ 3.9 & 64.4 $\pm$ 9.6 & \textbf{76.3} $\pm$ \textbf{5.4} & 67.9 $\pm$ 6.0 \\
        FedRep\\
                20& 81.1 $\pm$ 8.1 & \textbf{99.1} $\pm$ \textbf{0.0} & \textbf{99.1} $\pm$ \textbf{0.0} & 98.2 $\pm$ 1.3 && 75.6 $\pm$ 9.3 & \textbf{87.5} $\pm$ \textbf{4.5} & 81.4 $\pm$ 8.6 & 85.3 $\pm$ 7.3 \\
        40& 88.8 $\pm$ 10.4 & \textbf{98.9} $\pm$ \textbf{0.1} & 93.3 $\pm$ 7.1 & 96.7 $\pm$ 4.5 && 78.0 $\pm$ 6.3 & \textbf{88.0} $\pm$ \textbf{4.3} & 78.9 $\pm$ 7.9 & 76.7 $\pm$ 5.6 \\
        100& 93.0 $\pm$ 3.9 & \textbf{98.6} $\pm$ \textbf{0.1} & 98.4 $\pm$ 0.1 & 98.5 $\pm$ 0.1 && 86.0 $\pm$ 4.8 & \textbf{91.6} $\pm$ \textbf{3.1} & 89.1 $\pm$ 5.0 & 86.3 $\pm$ 4.9 \\
        FedPer\\
                20& 94.3 $\pm$ 4.3 & \textbf{99.5} $\pm$ \textbf{0.0} & \textbf{99.5} $\pm$ \textbf{0.0} & 99.3 $\pm$ 0.3 && 90.5 $\pm$ 2.4 & 92.7 $\pm$ 1.5 & 93.0 $\pm$ 4.3 & \textbf{93.8} $\pm$ \textbf{2.9} \\
        40& 94.7 $\pm$ 7.6 & \textbf{99.2} $\pm$ \textbf{0.1} & 99.1 $\pm$ 0.2 & 97.9 $\pm$ 2.7 && \textbf{92.3} $\pm$ \textbf{1.3} & 90.2 $\pm$ 4.7 & 87.7 $\pm$ 4.1 & 89.2 $\pm$ 2.3 \\
        100& 98.1 $\pm$ 0.1 & \textbf{98.9} $\pm$ \textbf{0.0} & 98.8 $\pm$ 0.2 & \textbf{98.9} $\pm$ \textbf{0.0} && \textbf{96.6} $\pm$ \textbf{0.9} & \textbf{96.6} $\pm$ \textbf{1.6} & 92.1 $\pm$ 3.3 & 90.2 $\pm$ 4.9 \\
                \cmidrule{3-9}
                Average Uplift& - & \textbf{26.4 $\pm$ 27.5} & 24.4 $\pm$ 26.5& 24.4 $\pm$ 25.6&& - & \textbf{12.7  $\pm$ 14.6} & 8.7  $\pm$ 12.7& 7.2  $\pm$ 11.4 \\
                \midrule
                \rowcolor{Gray} \multicolumn{10}{c}{             CIFAR10} \\\cmidrule{3-9}
        FedAvg\\
                20& 22.0 $\pm$ 2.6 & \textbf{75.1} $\pm$ \textbf{6.2} & 42.6 $\pm$ 4.5 & 52.2 $\pm$ 8.8 && 23.5 $\pm$ 6.9 & \textbf{71.4} $\pm$ \textbf{9.7} & 42.5 $\pm$ 4.7 & 49.7 $\pm$ 4.7 \\
        40& 26.1 $\pm$ 7.1 & \textbf{65.9} $\pm$ \textbf{7.1} & 36.7 $\pm$ 18.3 & 48.8 $\pm$ 8.3 && 26.6 $\pm$ 5.1 & \textbf{73.4} $\pm$ \textbf{15.9} & 36.3 $\pm$ 4.5 & 32.3 $\pm$ 11.6 \\
        100& 26.4 $\pm$ 4.3 & \textbf{68.0} $\pm$ \textbf{5.1} & 37.4 $\pm$ 11.4 & 39.8 $\pm$ 8.0 && 27.5 $\pm$ 2.0 & \textbf{54.6} $\pm$ \textbf{10.1} & 27.6 $\pm$ 4.1 & 29.0 $\pm$ 3.8 \\
        FedRep\\
                20& 81.8 $\pm$ 1.8 & \textbf{88.1} $\pm$ \textbf{2.0} & 84.4 $\pm$ 0.5 & 85.3 $\pm$ 0.5 && 85.3 $\pm$ 2.0 & \textbf{90.7} $\pm$ \textbf{2.5} & 87.9 $\pm$ 2.0 & 88.1 $\pm$ 1.4 \\
        40& 80.3 $\pm$ 0.8 & \textbf{83.7} $\pm$ \textbf{2.0} & 81.0 $\pm$ 2.1 & 81.6 $\pm$ 1.7 && 84.1 $\pm$ 0.8 & \textbf{93.6} $\pm$ \textbf{2.9} & 84.8 $\pm$ 1.7 & 84.3 $\pm$ 0.5 \\
        100& 75.0 $\pm$ 0.9 & \textbf{79.4} $\pm$ \textbf{2.3} & 75.2 $\pm$ 2.4 & 75.4 $\pm$ 1.5 && 77.9 $\pm$ 1.4 & \textbf{91.4} $\pm$ \textbf{2.0} & 77.8 $\pm$ 1.7 & 79.0 $\pm$ 1.1 \\
        FedPer\\
                20& 85.4 $\pm$ 2.3 & \textbf{91.0} $\pm$ \textbf{1.9} & 87.2 $\pm$ 0.5 & 87.8 $\pm$ 0.9 && 88.7 $\pm$ 1.7 & \textbf{92.3} $\pm$ \textbf{1.8} & 89.8 $\pm$ 2.0 & 90.1 $\pm$ 1.5 \\
        40& 85.9 $\pm$ 0.8 & \textbf{87.2} $\pm$ \textbf{2.2} & 82.7 $\pm$ 2.5 & 84.3 $\pm$ 1.9 && 88.1 $\pm$ 0.7 & \textbf{94.8} $\pm$ \textbf{2.6} & 86.0 $\pm$ 2.3 & 84.9 $\pm$ 3.3 \\
        100& 82.2 $\pm$ 0.4 & \textbf{85.1} $\pm$ \textbf{1.8} & 80.3 $\pm$ 2.0 & 80.9 $\pm$ 1.7 && 85.1 $\pm$ 0.6 & \textbf{94.0} $\pm$ \textbf{1.4} & 82.0 $\pm$ 2.4 & 83.0 $\pm$ 1.1 \\ \cmidrule{3-9}
        Average Uplift& - & \textbf{17.6 $\pm$ 19.6} & 4.7 $\pm$ 7.3& 7.9 $\pm$ 10.9&& - & \textbf{18.8  $\pm$ 16.6} & 3.1  $\pm$ 6.6& 3.7  $\pm$ 8.3    \\  \bottomrule \end{tabular}
      \vspace{-0.5cm}  
      \end{table*}

\section{Conclusion}
\label{sec:conclusion}
In this paper, we presented a principled approach for computing the Wasserstein 
distance between two distributions in a federated manner. 
Our proposed algorithm, called \ours, leverages the geometric 
properties of the Wasserstein distance and associated geodesics 
to estimate the distance while respecting the privacy of the 
samples stored on different devices. We established the convergence
 properties of \ours and provided empirical evidence of
its practical effectiveness through simulations on various problems, 
including dataset distance and coreset computation.
Our approach shows potential applications in the fields of machine learning 
and privacy-preserving data analysis, where computing distances  for 
distributed data is a fundamental task. 


\bibliographystyle{iclr2024_conference}

\newpage
\begin{appendices}
   \appendix

\setcounter{theorem}{0}

\section{Property of the approximating interpolating measure}
\label{sec:proof_approx}
\begin{theorem} 
Assume that $\mu$ and $\xi^{(k)}$ are two discrete distributions with the same number of samples $n$ and uniform weights.,
Then for any $t$, the approximating interpolating measure given by equation
\Cref{eq:approxinterpolant} is equal to the exact one \Cref{eq:interpolant}.
\end{theorem}
\begin{proof}
Remind that the approximating interpolating measure is defined as
\begin{equation}
  \xi_t = \frac{1}{n}\sum_{i=1}^n \delta_{(1-t)x_i + t n(\P^\star \mathbf{X}^\prime)_i}  
  \end{equation}
  whereas the exact interpolating measure is defined as
  \begin{equation}
    \mu_t = \sum_{i,j}^{n,m} \P_{i,j}^\star \delta_{(1-t)x_i + t x_j^\prime}
\end{equation}
where $\P^\star$ is the optimal transportation plan between $\xi$ and $\xi^\prime$, $x_i$ and
 $x_j^\prime$ are the samples from these distributions and $\mathbf{X}^\prime$ is the matrix
    of samples from $\xi^\prime$. 
Because $\mu$ and $\xi^{(k)}$ have the same number of samples $n$ and uniform weights, $\P^\star$
is a weighted (by $1/n$) permutation matrix \cite{peyre2019computational}. Let us denote by $\sigma$ the 
permutation associated to $\P^\star$.
Then, for the approximation, we have
\begin{align*}
  \xi_t &= \frac{1}{n}\sum_{i=1}^n \delta_{(1-t)x_i + t n(\P^\star \mathbf{X}^\prime)_i} \\
    &= \frac{1}{n}\sum_{i=1}^n \delta_{(1-t)x_i + t x_{\sigma(i)}^\prime} \\
  &= \sum_{i=1}^n \frac{1}{n} \delta_{(1-t)x_i + t x_{\sigma(i)}^\prime} \\
  & = \mu_t
  \end{align*}
where the last equality comes from the fact that for each row $i$, $P_{i,j}^\star$ is non-zero
only for $\sigma(i)$ column and $P_{i,\sigma(i)}^\star = 1/n$.
\end{proof}

\section{Proof of \Cref{thm:convergence}}
\label{sec:proof1}
\begin{theorem} 
    Let $\mu$ and $\nu$ be two measures in $\mathscr{P}_p(X)$. For $k \in \mathbb{N}$, let $\xi_{\mu}^{(k)}$, $\xi_{\nu}^{(k)}$
    and $\xi^{(k)}$ be interpolating measures computed at iteration $k$ as defined
    in \Cref{algo:fedOT}.
    Define 
    $$A^{(k)} = \mathcal{W}_p(\mu, \xi_{\mu}^{(k)}) + \mathcal{W}_p(\xi_{\mu}^{(k)}, 
    \xi^{(k)})+ \mathcal{W}_p(\xi^{(k)}, \xi_{\nu}^{(k)}) + \mathcal{W}_p(\xi_{\nu}^{(k)}, \nu)
    $$ 
    Then, the sequence $(A^{(k)})$ is non-increasing and converges to $\mathcal{W}_p(\mu, \nu)$.
\end{theorem}
\begin{proof}
First, remind that $\xi_{\mu}^{(k)}$ and $\xi_{\nu}^{(k)}$ are the interpolating measures between $\mu$ and
 $\xi^{(k-1)}$ and between $\xi^{(k-1)}$ and $\nu$ respectively, as defined in \Cref{algo:fedOT}. Likewise, 
 $\xi_{\mu}^{(k+1)}$ and $\xi_{\nu}^{(k+1)}$ are  interpolating measures between $\mu$ and 
 $\xi^{(k)}$ and between $\xi^{(k)}$ and $\nu$ respectively. Hence, we have
$$
    \mathcal{W}_p(\mu, \xi_{\mu}^{(k+1)}) + \mathcal{W}_p(\xi_{\mu}^{(k+1)}, \xi^{(k)} )  \leq
    \mathcal{W}_p(\mu, \xi_{\mu}^{(k)}) + \mathcal{W}_p(\xi_{\mu}^{(k)}, \xi^{(k)} )
$$
 and 
 $$
 \mathcal{W}_p(\nu, \xi_{\nu}^{(k+1)}) + \mathcal{W}_p(\xi_{\nu}^{(k+1)}, \xi^{(k)} )  \leq
 \mathcal{W}_p(\nu, \xi_{\nu}^{(k)}) + \mathcal{W}_p(\xi_{\nu}^{(k)}, \xi^{(k)} )
$$
These two inequalities lead to,
\begin{align*}
    &\mathcal{W}_p(\mu, \xi_{\mu}^{(k+1)}) + \mathcal{W}_p(\xi_{\mu}^{(k+1)}, \xi^{(k)} ) + 
    \mathcal{W}_p(\nu, \xi_{\nu}^{(k+1)}) + \mathcal{W}_p(\xi_{\nu}^{(k+1)}, \xi^{(k)} ) \\
    &\leq \mathcal{W}_p(\mu, \xi_{\mu}^{(k)}) + \mathcal{W}_p(\xi_{\mu}^{(k)}, \xi^{(k)} ) + 
    \mathcal{W}_p(\nu, \xi_{\nu}^{(k)}) + \mathcal{W}_p(\xi_{\nu}^{(k)}, \xi^{(k)} )
 \end{align*}
 Besides, since $\xi^{(k+1)}$ is an interpolating measure between $\xi_{\mu}^{(k+1)}$
  and $\xi_{\nu}^{(k+1)}$,
  we have 
  $$
    \mathcal{W}_p(\xi_{\mu}^{(k+1)}, \xi^{(k+1)} ) + \mathcal{W}_p( \xi^{(k+1)}, \xi_{\nu}^{(k+1)} ) \leq
    \mathcal{W}_p(\xi_{\mu}^{(k+1)}, \xi^{(k)} ) + \mathcal{W}_p(\xi^{(k)}, \xi_{\nu}^{(k+1)})
  $$
  and 
  $$
  A^{(k+1)} \leq A^{(k)}
  $$
    Hence, the sequence $(A^{(k)})$ is non-increasing. Additionally, by the triangle inequality, we have for any $k \in \mathbb{N}$,
    $$
    \mathcal{W}_p(\mu,\nu) \leq A^{(k)}
    $$ 
    We conclude by using the monotone convergence theorem: since $(A^{(k)})$ is non-increasing and bounded sequence below, then it converges to its infimum.

        We now justify why the limit of $(A^{(k)})$ is $\mathcal{W}_p(\mu, \nu)$. At convergence,
    we have reached a stationary point in the $(A^{(k)})$,
    \begin{align*}
      \lim_{k\to+\infty} A^{(k)} &= \mathcal{W}_p(\mu, \xi_{\mu}^{(\infty)}) + 
      \mathcal{W}_p(\xi_{\mu}^{(\infty)}, \xi^{(\infty)}) + \mathcal{W}_p(\xi^{(\infty)}, \xi_{\nu}^{(\infty)}) +
      \mathcal{W}_p(\xi_{\nu}^{(\infty)}, \nu) 
      \end{align*}
    however there are an 
    infinite number of triplets $(\xi_{\mu}^{(k)}, \xi_{\nu}^{(k)}, \xi^{(k)})$ that 
    allow to reach this value by the nature of the algorithm. By definition,
    $\xi_{\mu}^{(\infty)}$ and $\xi_{\nu}^{(\infty)}$ are interpolating measures between $\mu$
    and $\xi^{(\infty)}$ and between $\xi^{(\infty)}$ and $\nu$ respectively. Hence, by choosing
    $t=0$ and $t=1$ in the definition of the interpolating measure, we have $\xi_{\mu}^{(\infty)} = \mu$
    and $\xi_{\nu}^{(\infty)} = \nu$. Therefore, $\xi^{(\infty)}$ is also an interpolating measure between
    $\mu$ and $\nu$ as belonging to the geodesic between $\mu$ and $\nu$. Then, we have
    \begin{align*}
      \lim_{k\to+\infty} A^{(k)} &= \mathcal{W}_p(\mu, \xi_{\mu}^{(\infty)}) + 
      \mathcal{W}_p(\xi_{\mu}^{(\infty)}, \xi^{(\infty)}) + \mathcal{W}_p(\xi^{(\infty)}, \xi_{\nu}^{(\infty)}) +
      \mathcal{W}_p(\xi_{\nu}^{(\infty)}, \nu) \\
      &= \mathcal{W}_p(\mu, \xi^{(\infty)}) + \mathcal{W}_p(\xi^{(\infty)}, \nu) \\
      &= \mathcal{W}_p(\mu, \nu)
      \end{align*}
      where the first equality results from the fact that  $\xi_{\mu}^{(\infty)}$ and $\xi_{\nu}^{(\infty)}$ are interpolating measures between $\mu$
      and $\xi^{(\infty)}$ and between $\xi^{(\infty)}$ and $\nu$ respectively and the second equality is obtained  from the fact that
      $\xi^{(\infty)}$ is also an interpolating measure between
      $\mu$ and $\nu$ as belonging to the geodesic between $\mu$ and $\nu$.
\end{proof}

Let us note that the convergence of the algorithm requires choosing $\xi_{\mu}^{(\infty)} = \mu$
and $\xi_{\nu}^{(\infty)} = \nu$ thus leaking the client distributions to the server.
However, this is just a theoretical argument that we use just to show that the algorithm
converges so that    $\lim_{k\to+\infty} A^{(k)} = \mathcal{W}_p(\mu, \nu)$. In practice,
we avoid setting $t=0$ and $t=1$ in the definition of the interpolating measures.

\section{Convergence rate of the algorithm for Gaussian distributions with
same covariance}
\label{sec:proof_convergence}

In this section, we show that when $\mu$ and
$\nu$ are Gaussians after one iteration, we can infer $W(\mu,nu)$ and 
the sequence of iterates $(\xi^{(k)})$ obtained for $t=0.5$ converges to the $\xi^\star$ 
the interpolating measure between $\mu$ and $\nu$ for $t=0.5$

\begin{theorem}
  Assume that $\mu$, $\nu$ and $\xi^{(0)}$ are three Gaussian distributions with the same covariance matrix $\Sigma$
  ie $\mu \sim \mathcal{N}(\m_\mu, \Sigma)$, $\nu \sim \mathcal{N}(\m_\nu, \Sigma)$
  and $\xi^{(0)} \sim \mathcal{N}(\m_{\xi^{(0)}}, \Sigma)$. Further assume that we are not
  in the trivial case where $\m_\mu$, $\m_\nu$, and $\m_{\xi^{(0)}}$ are aligned.
  Applying our algorithm \Cref{algo:fedOT} with $t=0.5$ and the squared Euclidean cost, we have the following properties:
  \begin{enumerate}
    \item all interpolating measures $\xi_{\mu}^{(k)}$,$\xi_{\nu}^{(k)}$, $\xi^{(k)}$ are isotropic Gaussian distributions with the same covariance matrix $\Sigma$
    \item for any $k \geq 1$, $\mathcal{W}_2(\mu,\nu) = \| \m_\mu - \m_\nu\|_2 = 2 \| \m_{\xi_\mu^{(k)}} - \m_{\xi_\nu^{(k)}}\|_2 $
    \item $\mathcal{W}_2(\xi^{(k)},\xi^{\star}) = \frac{1}{2} \mathcal{W}_2(\xi^{(k-1)},\xi^{\star})$ 
    \item $\mathcal{W}_2(\mu,\xi^{(k)}) + \mathcal{W}_2(\xi^{(k)},\nu) - \mathcal{W}_2(\mu,\nu)\leq \frac{1}{2^{k-1}} \mathcal{W}_2(\xi^{(0)},\xi^{(\star)})$
  \end{enumerate}

\end{theorem}
\begin{proof}
The first point comes from the fact that Wasserstein barycenter of Gaussians are Gaussians 
\cite{agueh2011barycenters, peyre2019computational}. For isotropic Gaussians with same covariance,
the covariance matrice of the barycenter remains unchanged while the mean is the barycenter
mean. So, in our case, the interpolating measure with $t=0.5$ \emph{i.e} the uniform barycenter
of two measures, say $\mu$ and $\xi^{(k-1)}$, is $\xi_\mu^{(k)} \sim \mathcal{N}(\m_{\xi_\mu^{(k)}},
\Sigma)$, where $ \m_{\xi_\mu^{(k)}} = \frac{1}{2}(\m_\mu + \m_{\xi^{(k-1)}})$.
The consequence of this first point of the theorem is that since we are going to deal 
with same covariance Gaussian distributions, then the Wasserstein distance between any pair 
of measures involved in our algorithm only depends on the Euclidean distance of their means and
we will use interchangeably the Euclidean distance and the Wasserstein distance. 
~\\
~\\
The second point is proven by using geometrical arguments in the plane $(P)$ in which the 
three points, for $k \geq 1$, $\m_\mu$, $\m_\nu$, $\m_{\xi^{(k-1)}}$ lie (note that based on our assumption, this plane
always exists). 
By definition of $\xi_\mu^{(k)}$ and $\xi_\nu^{(k)}$ and given the above point, we have
$$
\m_{\xi_\mu^{(k)}} = \frac{1}{2}(\m_\mu + \m_{\xi^{(k-1)}})\quad \text{and} \quad 
\m_{\xi_\nu^{(k)}} = \frac{1}{2}(\m_\nu + \m_{\xi^{(k-1)}})
$$ 
By using the intercept theorem, since $t=\frac{1}{2}$, in the plane $(P)$, the segment
$[\m_{\xi_\mu^{(k)}}, \m_{\xi_\nu^{(k)}}]$ is parallel to the segment
$[\m_{\mu}, \m_{\nu}]$ and we have :
$$
\frac{1}{2} =  \frac{\|\m_{\xi_\mu^{(k)}} - \m_{\xi{(k-1)}}\|_2}{\|\m_{\mu} - \m_{\xi{(k-1)}}\|_2}
= \frac{\|\m_{\xi_\nu^{(k)}} - \m_{\xi{(k-1)}}\|_2}{\|\m_{\nu} - \m_{\xi{(k-1)}}\|_2} 
= = \frac{\|\m_{\xi_\nu^{(k)}} - \m_{\xi_\mu^{(k)}} \|_2}{\|\m_{\nu} - \m_{\mu}\|_2}  
$$
which gives us the second point. 
~\\
~\\
For the third point,  we are going to consider geometrical arguments similar as above. However,
we are going to first show that for a given $k$, the mid point, denoted as $\hat \xi^{(k)}$, between
$\xi^{(k-1)}$ and $\xi^{\star}$ is also  $\xi^{(k)}$ as defined by our algorithm.  

By definition, $\xi^{\star}$ is the mid point interpolating measure between $\mu$ and $\nu$,
whose mean is $\frac{1}{2}(\m_\mu + \m_\nu)$. Since $\hat \xi^{(k)}$ and $\xi_\mu^{(k)}$ are
respectively the mid point measure between $\xi^{(k-1)}$ and $\xi^{\star}$ and $\mu$ and $\xi^{(k-1)}$,
we can apply the intercept theorem in the appropriate plane and get 
$$
W_2(\hat \xi^{(k)}, \xi_\mu^{(k)}) = \frac{1}{2}W_2(\mu, \xi^\star)
$$  
Using a similar reasoning using $\nu$, we get
$$
W_2(\hat \xi^{(k)}, \xi_\nu^{(k)}) = \frac{1}{2}W_2(\nu, \xi^\star)
$$  

Summing these two equations, we obtain
$$
W_2(\hat \xi^{(k)}, \xi_\mu^{(k)}) + 
W_2(\hat \xi^{(k)}, \xi_\nu^{(k)}) = \frac{1}{2} W_2(\mu, \xi^\star) + \frac{1}{2} W_2(\nu, \xi^\star)
= \frac{1}{2} W_2(\mu,\nu) = W_2(\xi_\mu^{(k)},\xi_\nu^{(k)}) $$
where the second equality comes from the fact that $\xi^\star$ is an interpolant measure of
$\mu$ and $\nu$, while the last equality comes from the second point of the theorem.

Hence, since we have $W_2(\hat \xi^{(k)}, \xi_\mu^{(k)}) + 
W_2(\hat \xi^{(k)}, \xi_\nu^{(k)}) = W_2(\xi_\mu^{(k)},\xi_\nu^{(k)})$, it also mean
than 
$$
\hat \xi^{(k)} \in \argmin_\xi \frac{1}{2} W_2(\xi_\mu^{(k)}, \xi) + 
\frac{1}{2} W_2(\xi, \xi_\nu^{(k)})
$$
and $\hat \xi^{(k)}$ is also the midpoint interpolating measure between $\xi_\mu^{(k)}$
and $\xi_\nu^{(k)}$. 

Then, applying the intercept theorem with $\xi^{(k-1)}$, $\xi^{(k)}$, $\xi^{\star}$, 
$\mu$ and $\xi_{\mu}^{(k)}$, we obtain the desired result
$$
\frac{1}{2}W_2(\xi^{(k-1)},\xi^{\star}) =  W_2(\xi^{(k)}, \xi^{\star})
$$ 

Finally, given all the above, it is simple to show the convergence rate of the algorithm using
simple triangle inequalities.
\begin{eqnarray}
  \mathcal{W}_2(\mu,\xi^{(k)}) + \mathcal{W}_2(\xi^{(k)},\nu) - \mathcal{W}_2(\mu,\nu) &\leq& 
  \mathcal{W}_2(\mu,\xi^\star) + \mathcal{W}_2(\xi^{\star}, \xi^{(k)}) + \mathcal{W}_2(, \xi^{(k)}, \xi^{\star}) + \mathcal{W}_2(\xi^{\star},\nu) - \mathcal{W}_2(\mu,\nu) \nonumber \\
    &=& 2 \mathcal{W}_2(\xi^{(k)},\xi^{\star}) \nonumber \\
  &=& \frac{1}{2^{k-1}} \mathcal{W}_2(\xi^{(0)},\xi^{\star}) \nonumber
\end{eqnarray}

\end{proof}

\begin{figure*}
  \centering
\begin{tikzpicture}
    \coordinate (A) at (0,0);
  \coordinate (B) at (6,0);
  \coordinate (C) at (2,3);

    \draw (A) -- (B) -- (C) -- cycle;

    \node[below left] at (A) {$\mu$};
  \node[below right] at (B) {$\nu$};
  \node[above] at (C) {$\xi^{(k-1)}$};

    \coordinate (M1) at ($(A)!0.5!(C)$);
  \coordinate (M2) at ($(B)!0.5!(C)$);
  \coordinate (M3) at ($(B)!0.5!(A)$);
  \coordinate (M4) at ($(M1)!0.5!(M2)$);

  \node[left] at (M1) {$\xi_\mu^{(k)}$};
  \node[right] at (M2) {$\xi_\nu^{(k)}$};
  \node[below] at (M3) {$\xi^\star$};
  \node[above right] at (M4) {$\xi^{(k)}$};
  \node[below left] at (M4) {$\hat \xi^{(k)}$};

    \draw (M1) -- ($(M1)!1!(M2)$);

    \draw (M3) -- ($(M3)!1!(C)$);
\end{tikzpicture}
\caption{Illustration of the geometrical interpretation of the algorithm and its convergence
proof for Gaussian distributions with same covariance, based on the intercept theorem.}
\end{figure*}
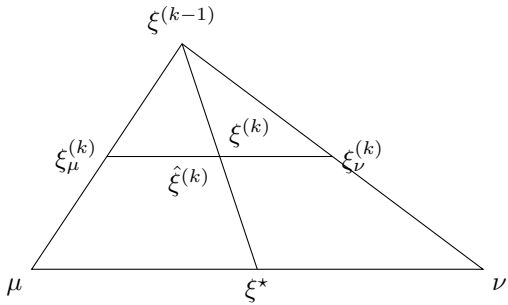

\section{Additional experiments}
\label{app:expe}

\subsection{Toy Analysis : the impact of approximating the interpolating measure}

We propose to analyze in this section the benefits and disadvantages of approximating the interpolating measure
instead of using the exact one as given in \Cref{eq:interpolant}.
For this purpose, we compare the running time and the accuracy of the exact Wasserstein distance, 
our exact FedWaD, and our approximate FedWaD for estimating the Wasserstein distance between 
two Gaussians distributions.  
The Gaussians are different means but same covariances so that the true
Wasserstein distance is known and equal to the Euclidean distance between the means.
We have considered two different settings ($d=2$ and $d=50$) of Gaussians dimensionality.
For the first case ($d=2$), we detail the results presented in the main paper.
Note that when the dimensionality of the Gaussians are set to $50$, we do not expect the Wasserstein distance nor \ours to 
provide a good estimation of the  closed form distance between these distributions,
due to the curse of dimensionality of the Wasserstein distance \citep{fournier2015rate}

As default parameter for our approximate FedWaD, we considered $20$ iterations and a support of size $10$, 
then we varied the number of samples $n$ from $10$ to $10000$. We have run experiments in different settings

\begin{itemize}
  \item we analyzed the impact of sample ratio between the two distributions, as this may impact the 
  support size of the approximating interpolating measure accross \ours iterations.
\item we made varying the support size of the approximating interpolating measure at fixed sample ratio.
\end{itemize}

Results have been averaged over $10$ runs.

\begin{figure}[t]
  ~\hfill~\includegraphics[width=7cm]{sample_ratio_time_2.pdf}~\hfill~
  \includegraphics[width=7cm]{sample_ratio_error_2.pdf}~\hfill~ \\ 
  ~\hfill~\includegraphics[width=7cm]{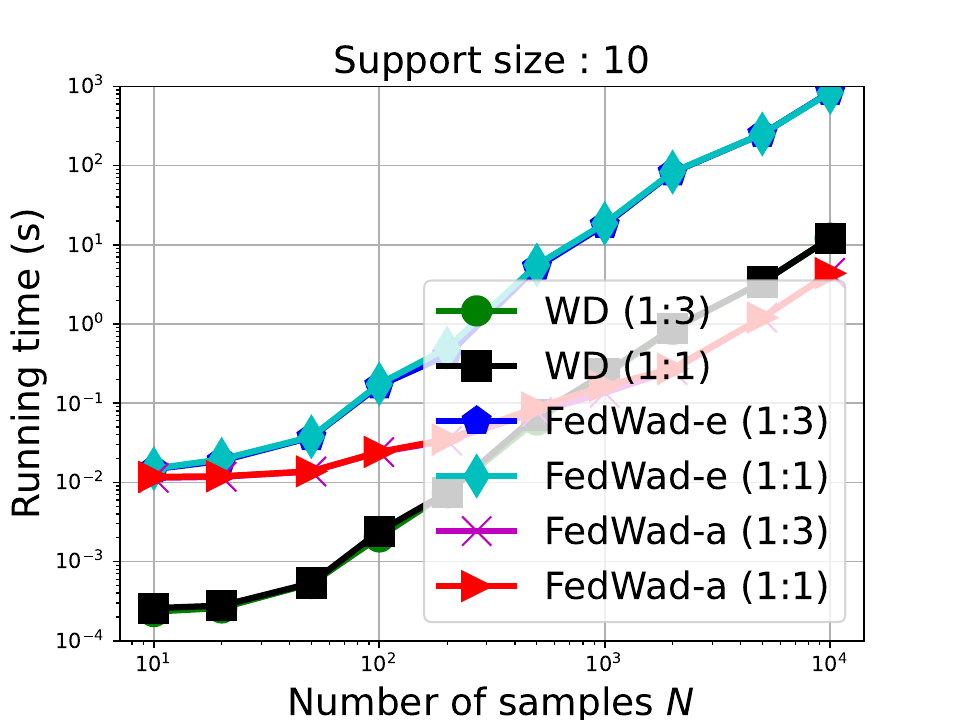}~\hfill~
  \includegraphics[width=7cm]{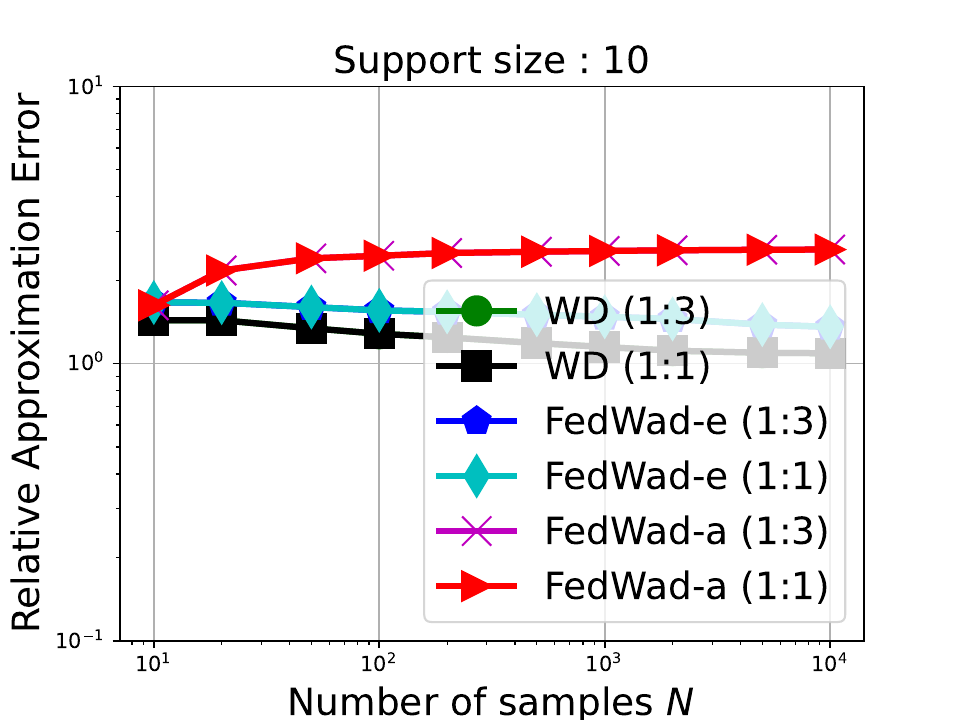}~\hfill~  
  \caption{For different sample ratios, (1:3) or (1:1), in the two distributions 
  we report the performance of the different models. For our approximated FedWaD, we have
  set the support size to $10$. (top) $d=2$ (bottom) $d=50$. (left) running time. (right) relative error.}
  \label{fig:sample_ratio_50}
\end{figure}

\begin{figure}[t]
~\hfill~\includegraphics[width=7cm]{support_time_2.pdf}~\hfill~
\includegraphics[width=7cm]{support_error_2.pdf}~\hfill~\\
~\hfill~\includegraphics[width=7cm]{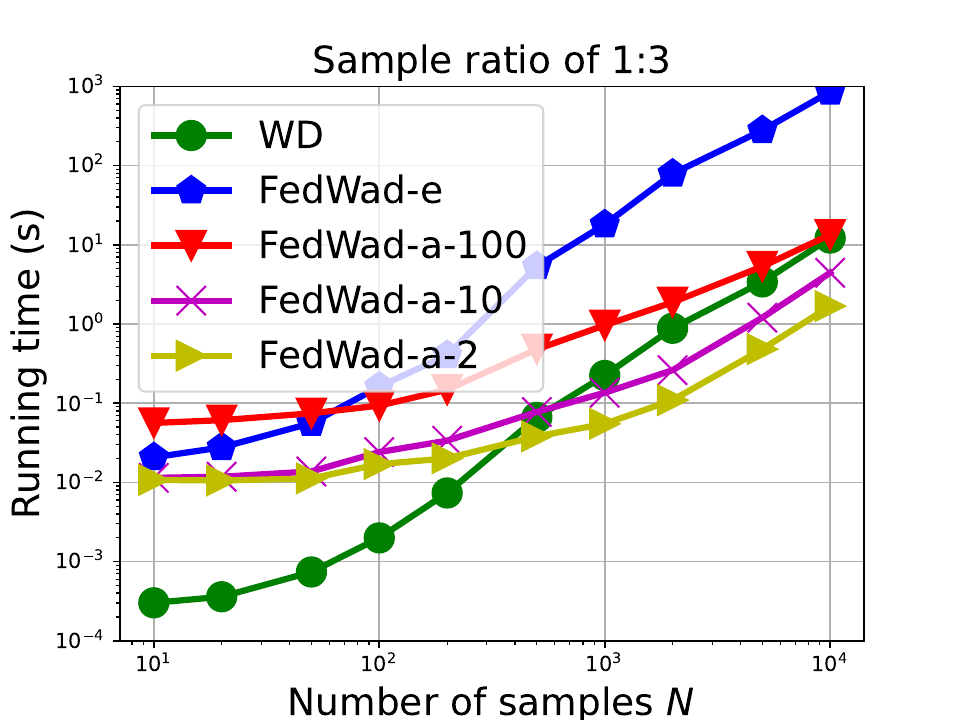}~\hfill~
\includegraphics[width=7cm]{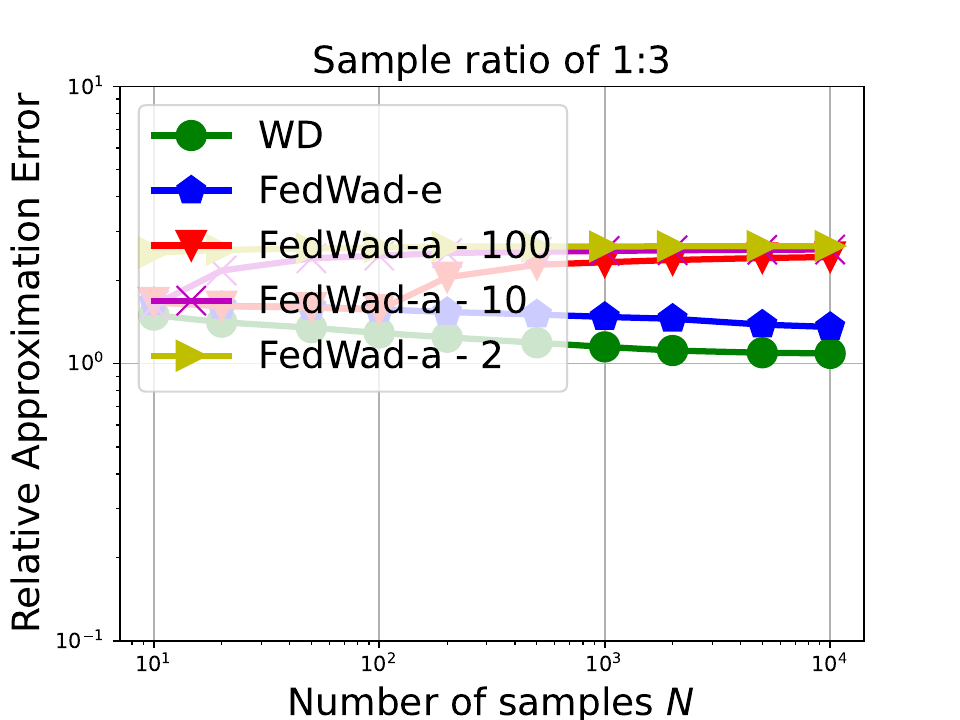}~\hfill~
  \caption{For increasing number of samples, we report 
  (top) $d=2$ (bottom) $d=50$. (left) Running time of the Wasserstein distance, our exact FedWaD and
   our approximate FedWaD. (right) the relative error of the different models : the
   computed Wasserstein distance, our exact FedWaD and the approximated FedWad with a support
   size of $10$ and $100$.  The first 
   distribution has a number of samples $N$ and the second ones $N/3$. }
  \label{fig:support_diff_50}
\end{figure}

\paragraph{Analyzing the impact of sample ratio}  
Given the setting with uniform weights, when the sample ratio is $1$, the optimal
plan is theoretically a scaled permutation matrix. Hence, the support size of the exact 
interpolating measure is expected, in theory, to be fixed and
equal to $N$. When the ratio of samples is different to $1$, the support size of the exact interpolating
may increase at each iteration of the algorithm and leads to a larger running time.
\Cref{fig:sample_ratio_50} - left panel - shows the running time of all compared methods as well as
their relative error - right panel - compared to the true Wasserstein distance. We note that
for $2d$ Gaussians, both the Wasserstein distance and our approximated FedWaD with support size of $10$ the running
time is increasing with a natural computational overhead for the 1:1 sample ratio (as we have more samples).
For the exact FedWaD, the behavior is different. the running time for the 1:3 sample ratio is
larger than the 1:1. This is due to  the optimal transportation plan $\P^\star$ not being exactly a scaled 
permutation matrix. As a result, the support size of the
interpolating measure increases with the number of samples, leading to computational overhead
for the method. For $50 d$ Gaussians, the differences in running time between the different sample
ratio are negligible.

In the case of $2d$ Gaussians (top row), For the relative error, for $N < 1000$, we note that all methods achieve similar errors. Numerical
errors start to appear for exact FedWaD and the Wasserstein distance for respectively $N \geq 1000$ and
$N \geq 5000$ depending on the sample ratio. Interestingly, the approximated FedWaD is robust to large
number of samples and achieves similar errors as for small number of samples. For higher
dimensions (bottom row), all the methods are not able to provide accurate estimation of the Wasserstein
distance and with the worst relative error for the approximated FedWaD with a support size of $10$. 
Nonetheless, we want to emphasize that despite this lack of accuracy, the approximated \ours can be
useful in high-dimension problems as we have shown for the other experiments.

\paragraph*{Analyzing the support size of approximated interpolating measure} 
\Cref{fig:support_diff_50} shows the running time and the relative error of the different methods
for a sample ratio of $1:3$ and when the support sizes of the approximating interpolating measure
are $2$,$10$ or $100$. We clearly remark the computational cost of a larger support size with
a benefit in terms of approximation error appearing mostly when $N \geq 1000$ and for small dimension
problems (top row).
For higher dimension problems (bottom row), we see again the benefit on running time of the 
approximated approach, yet with a larger approximation error.

\subsection{Details on Coreset and additonal results}

\paragraph{Experimental setting} We sampled $20000$ examples randomly from the \texttt{MNIST} dataset,
 and dispatched them at random on $100$ clients but 
 such that only a subset $K$ of the $10$ classes is present on each client.
We learn $10$ coresets over $1000$ epochs and at each epoch, we assume that only $10$ random clients
are available and can be used for computing \ours. For \ours, the support size of
the interpolating measure has been set to either $10$ or $100$ and the number of iteration in \ours to $20$.

\begin{figure}[t]    
  \begin{center}
  \begin{minipage}{0.6\textwidth} 
  \includegraphics[width=0.7\textwidth]{coreset-setting1-su8.pdf}
  \includegraphics[width=0.7\textwidth]{coreset-setting1-su2.pdf}
  \includegraphics[width=0.7\textwidth]{coreset-setting3-support.pdf}
  \end{minipage}
  \begin{minipage}{0.35\textwidth} 
  \hspace{-2.2cm}    \vspace{-0.4cm}
  ~\hfill
  \includegraphics[width=6cm]{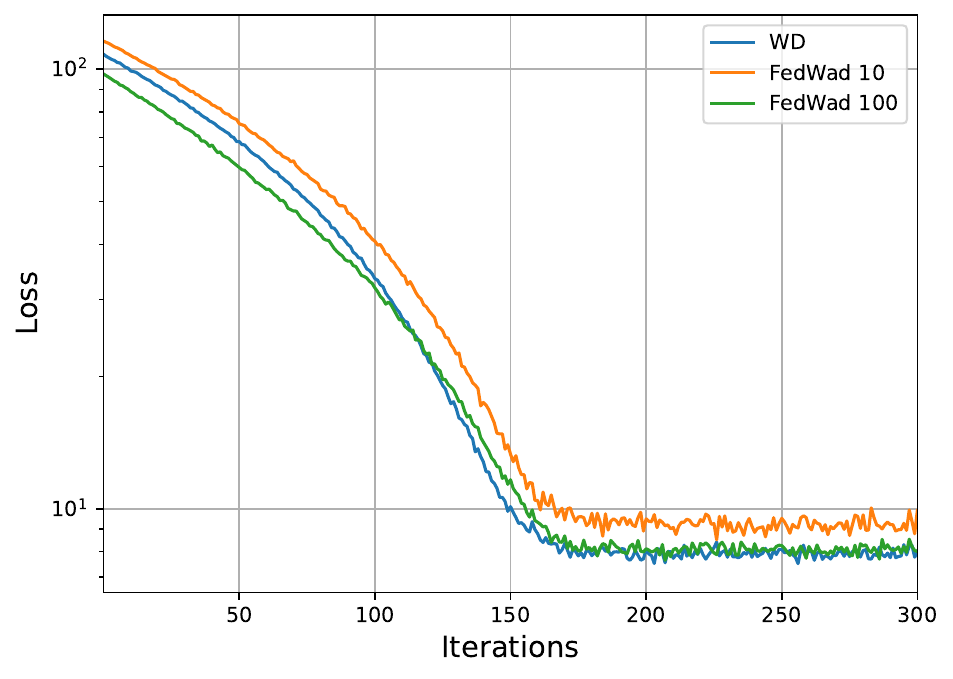}
\hfill~    
\end{minipage}
  \end{center}
  \caption{Examples of the $10$ coreset obtained with for each panel \emph{(top-row)} the exact Wasserstein
  and \emph{(bottow-row)} \ours for the \texttt{MNIST} dataset. Different panels correspond to different number of classes $K$ on each client: \emph{(top)} 
  $K=8$, \emph{(middle)} $K=2$, \emph{(bottom)} support of the interpolating measure
  varying from $10$ to $100$. As class
  diversity on each client increases, the coreset is less effective at capturing
  the $10$ modes of the dataset}
  \label{fig:coresetmnist}
\end{figure}

\begin{figure}[t]
  \

    \begin{center}
      \begin{minipage}{0.6\textwidth} 
      \includegraphics[width=0.7\textwidth]{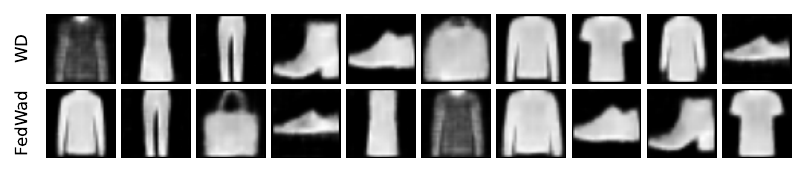}
      \includegraphics[width=0.7\textwidth]{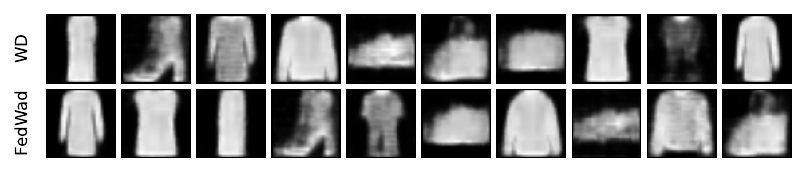}
      \includegraphics[width=0.7\textwidth]{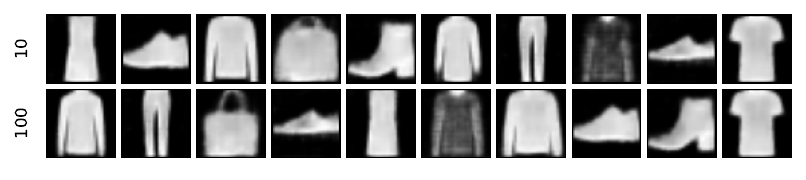}
      \end{minipage}
      \begin{minipage}{0.35\textwidth} 
      \hspace{-2.2cm}    \vspace{-0.4cm}
      ~\hfill
      \includegraphics[width=6cm]{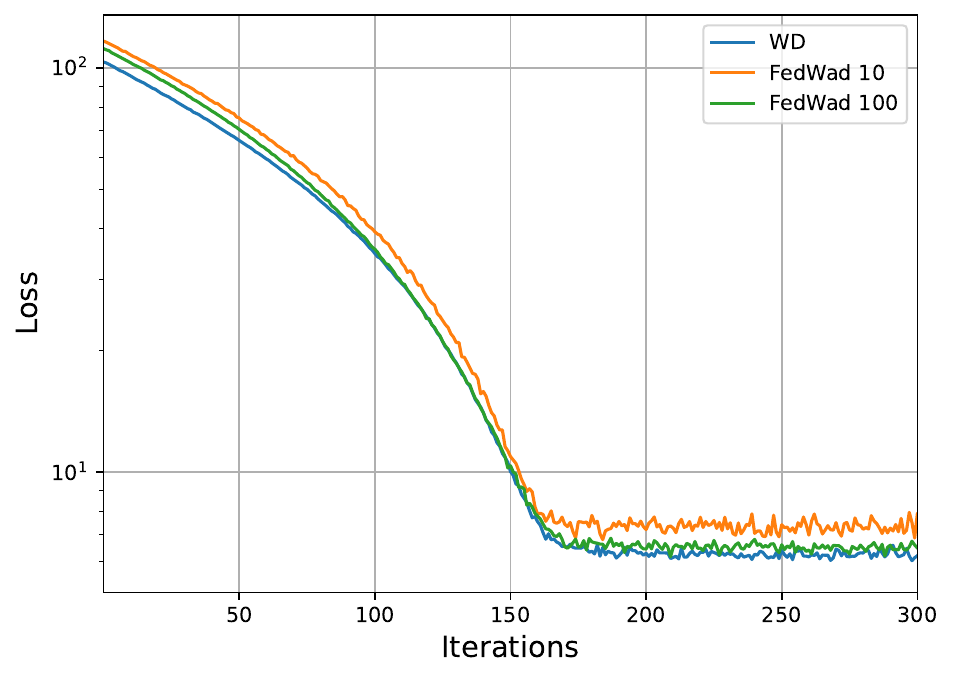}
  \hfill~    
  \end{minipage}
      \end{center}

  \caption{Examples of the $10$ coreset obtained with for each panel \emph{(top-row)} the exact Wasserstein
  and \emph{(bottow-row)}, our FedWaD for the \texttt{FashionMNIST} dataset. Different panels correspond to different number of classes $K$ on each client: \emph{(top)} 
  $K=8$, \emph{(middle)} $K=2$, \emph{(bottom)} support of the interpolating measure
  for $K=8$.}
  \label{fig:coresetfm}
\end{figure}

\begin{figure}[t]
  \begin{center}
  \includegraphics[width=6cm]{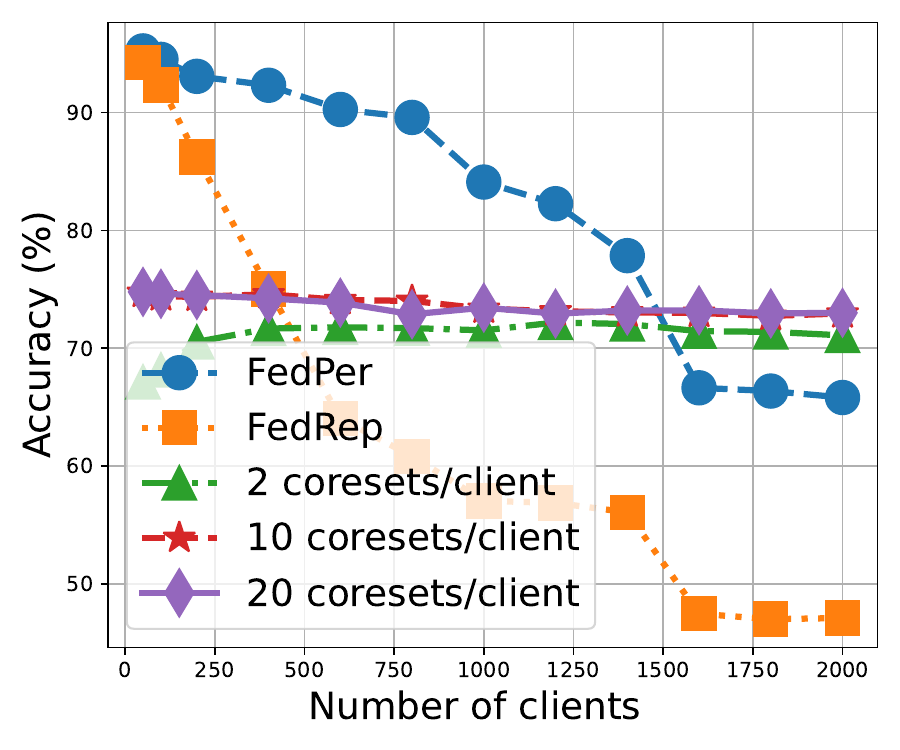} 
  \end{center}
  \caption{ FashionMNIST performance of a nearest neighbor classifier based on the coresets learnt from each client for varying number of clients and 
  number of coresets per clients We have compared to the performance of two personalized FL algorithms.}
  \label{fig:coreset_perf_fmnist}
\end{figure}

We have reproduced in here the same MNIST experiment (which results 
are reproduced in \Cref{fig:coresetmnist}) on coreset for the \texttt{FashionMNIST} 
dataset, and we can notice, in  \Cref{fig:coresetfm} that we obtain similar results as for the \texttt{MNIST} dataset.
When the number of shared classes $K$ is large enough, the coreset is not
able to capture the different modes in the dataset. And again, we remark that the
support size of the approximate interpolating measure has few impacts on the result.
For both datasets, the loss landscape of the coreset learning  reveals 
that our \ours-based approaches yield to a worse minimum than the exact Wasserstein distance, which is mostly due to the interpolating measure approximation.
\Cref*{fig:coreset_perf_fmnist} plots the performance of a nearest neighbor classifier based on the coresets learnt from each client for varying number of clients.
Results show that coreset-based approaches are competitive, especially for high number of clients, with personalized FL algorithms, 
which are known to be the best performing FL algorithms in practice.

\subsection{Details on Federated OTDD experiments}
\label{app:otdd}

\paragraph{Geometric dataset distances via federated Wasserstein distance.}

Transfer learning and domain adaptation are important ML paradigms, which aim 
at transferring knowledge across similar domains.  The main underlying concept in these approaches is the notion of distance or similarity between datasets. 
Transferring knowledge between comparable domains is typically simpler than between distant ones. In certain applications, it is relevant to find datasets from which one can transfer knowledge from without disclosing the target dataset.
This may be the case, for instance, in applications with low-resource clients storing sensitive data. In this case, the practitioner may want to find a dataset
similar enough to the client's dataset, in order to transfer knowledge from it.
In practice, a server would train a classifier on a dataset that is similar to the client dataset,
and the client would then use this classifier to perform inference on its own data.

In that context, our goal is to propose a distance between datasets that can be computed
in a federated way based on \ours. We leverage the distance proposed in 
\cite{alvarez2020geometric}, which is based on the Wasserstein distance between two labeled datasets
$\mathcal{D}$ and $\mathcal{D}^\prime$. The ground metric is defined by,
\begin{equation}
d_\mathcal{D}((x,y),(x^\prime,y^\prime)) \triangleq ( d(x,x^\prime) + \mathcal{W}_2^2(\alpha_y,\alpha_y^\prime) )^{1/2}
\end{equation}
where $d$ is a distance between two features $x$ and $x^\prime$, and $\alpha_y$ is the class-conditional distribution of $x$ given $y$. In order to reduce computational complexity, \cite{alvarez2020geometric} assume the class-conditionals are Gaussian, so that $\mathcal{W}_2$ boils down the $2$-Bures-Wasserstein distance, which is available in closed form:
\begin{equation}
 \mathcal{W}_2^2(\alpha_y, \alpha_{y^\prime}) = \|m_y - m_{y^\prime}\|_2^2  + \|\Sigma_y - \Sigma_{y^\prime}\|_F^2
\end{equation}
where $m_{z}$ and $\Sigma_{z}$ denote the mean and covariance of $\alpha_{z}$.

\ours needs vectorial representations of the data to compute intermediate measures.
The Bures-Wasserstein distance allows us to conveniently represent $\alpha_y$ as the concatenation of the mean $m_{y}$ and vectorized covariance $\Sigma_{y}$.
Hence, we can compute the distance between two datasets $\mathcal{D}$ and $\mathcal{D}^\prime$ by augmenting
each example from those datasets with the corresponding class-conditional mean and vectorized covariance, and
using the $\ell_2$ norm as the ground metric in the Wasserstein distance. One can eventually reduce the dimension the augmented representation by considering only the 
diagonal of the covariance matrix.

\subsection{Federated OTDD analysis}
\label{app:fedotddanalysis}

\begin{figure}[t]
    \begin{center}
        ~\hfill
    \includegraphics[width=0.4\textwidth]{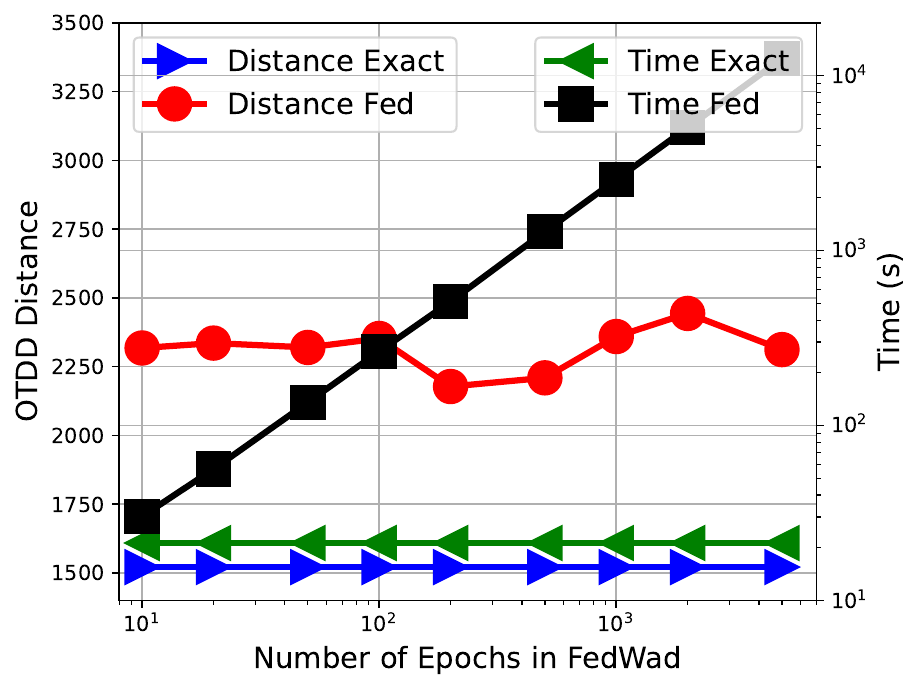}~\hfill~
    \includegraphics[width=0.4\textwidth]{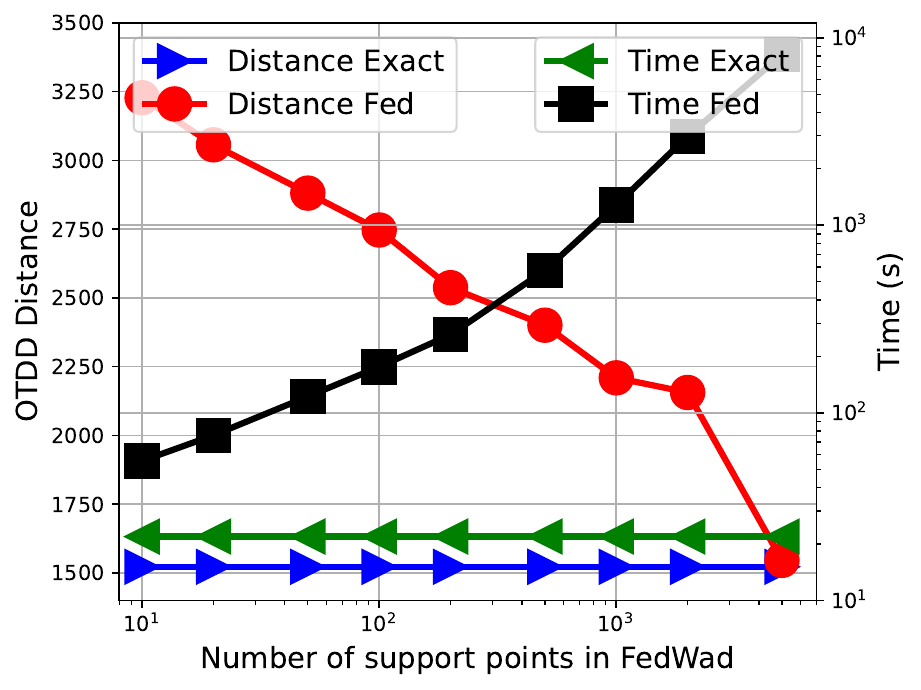}
    ~\hfill~    
\end{center}
    \caption{\ours and OTDD distances on \texttt{MNIST}-\texttt{KMNIST} and its running time against \emph{(left)} the number of epochs
     and \emph{(right)} the number of support points in the interpolating measure. For each plot, the left and right $y$-axis report the distance and the running time respectively.}     \label{fig:otddanalysis_apdx}
\end{figure}

To evaluate our procedure, we replicated the experiments of \cite{alvarez2020geometric} on the
dataset selection for transfer learning: given a source dataset,
the goal is to find a target one which is the most similar to the source. 
 We considered four real datasets, namely \texttt{MNIST}, \texttt{KMNIST},
 \texttt{USPS} and \texttt{FashionMNIST}. 
We first analyze the impact of two hyperparameters, the number of epochs and the number of 
support points in the interpolating measure, on the distance computation
between $5000$ samples from \texttt{MNIST} and \texttt{KMNIST}, \Cref{fig:otddanalysis_apdx}
 shows the evolution of the distance between 
\texttt{MNIST} and \texttt{KMNIST} as well as the running time
for varying values of hyperparameters. The number of epochs has a very small impact 
on the distance and using 
$10$ epochs suffices to get a reasonably accurate approximation of the distance. On the other hand, the number
of support point seems more critical, and we need at least $5000$ support points to obtain a very accurate
 approximation, although we have a nice linear convergence of the distance with respect to support size.

\begin{figure}[t]
  \begin{center}
      \includegraphics[width=0.4\textwidth]{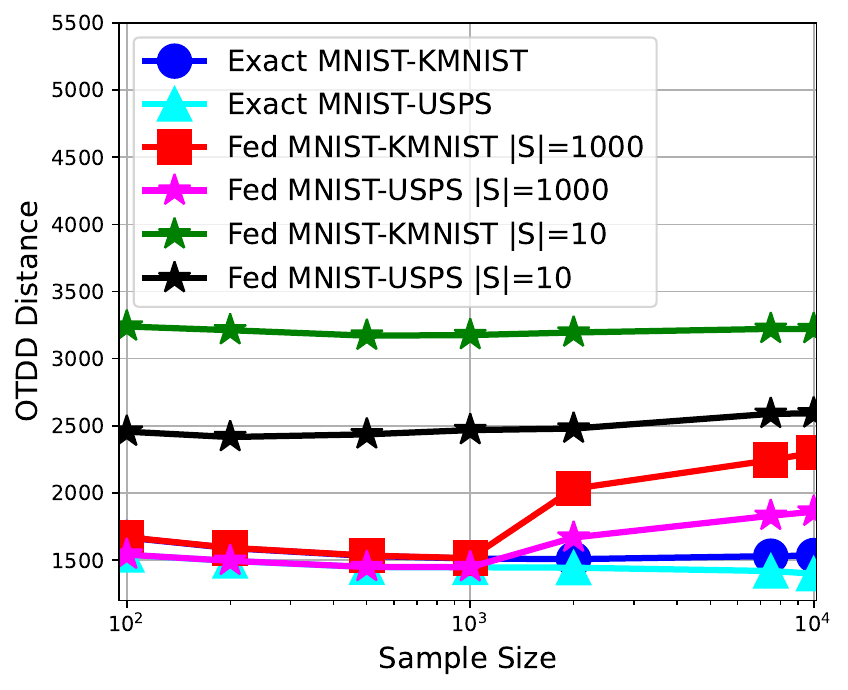}
      \includegraphics[width=0.4\textwidth]{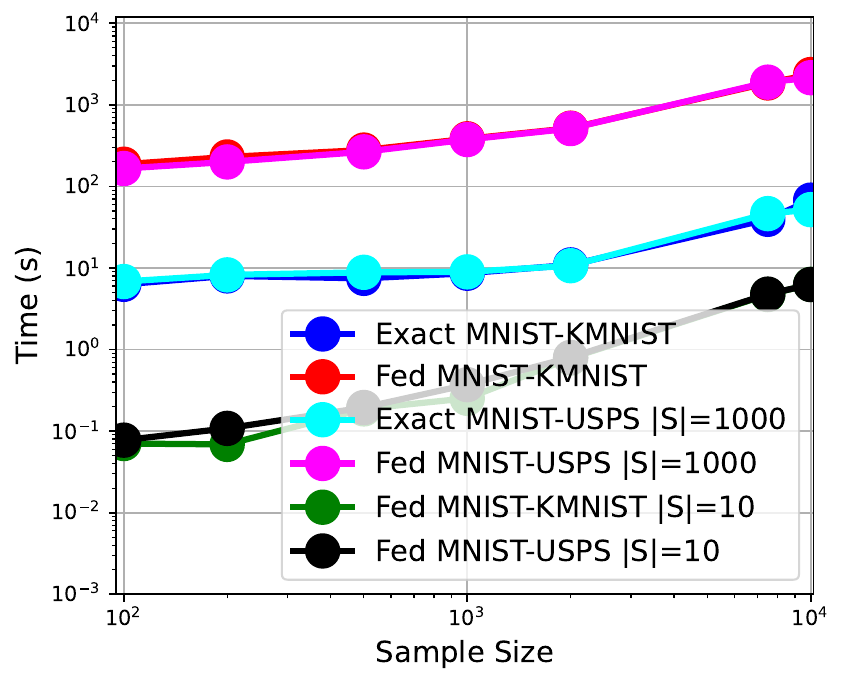}        
  \end{center}
  \caption{\emph{(left)} Distance and \emph{(right)} running time against the dataset size 
  for the \texttt{MNIST}-\texttt{KMNIST}  an \texttt{MNIST}-\texttt{USPS} distances, for varying number of support points $|S|$}
  \label{fig:otddsample_size_apdx}
\end{figure}

We  also analyzed the impact of the dataset size on the distance computation and running time: \Cref{fig:otddsample_size_apdx} shows the evolution of the distance and the running time with respect to the
the sample size in the two distributions. We note that the order relation
is preserved between the two distances for all possible range of sample size.
Another interesting observation is that as long as the sample size is smaller than the support size
of the interpolating measure,
\ours provides an accurate estimation of the distance. When the sample size is larger
 then the distance is overestimated. This is due to a less accurate estimation of an exact interpolating
 measure (which is supported on $2n+1$ points). Regarding computational efficiency, we observe that
 for small support size of the interpolating measure, the running time increases at the same
 rate as the sample size, whereas for larger support size, the running time increases $10$-fold
 for an $100$-fold increase in sample size.

\subsection{Boosting FL methods}
\label{sec:boosting_apdx}

\begin{figure}[t]~\\
~\hfill~\includegraphics[width=6.5cm]{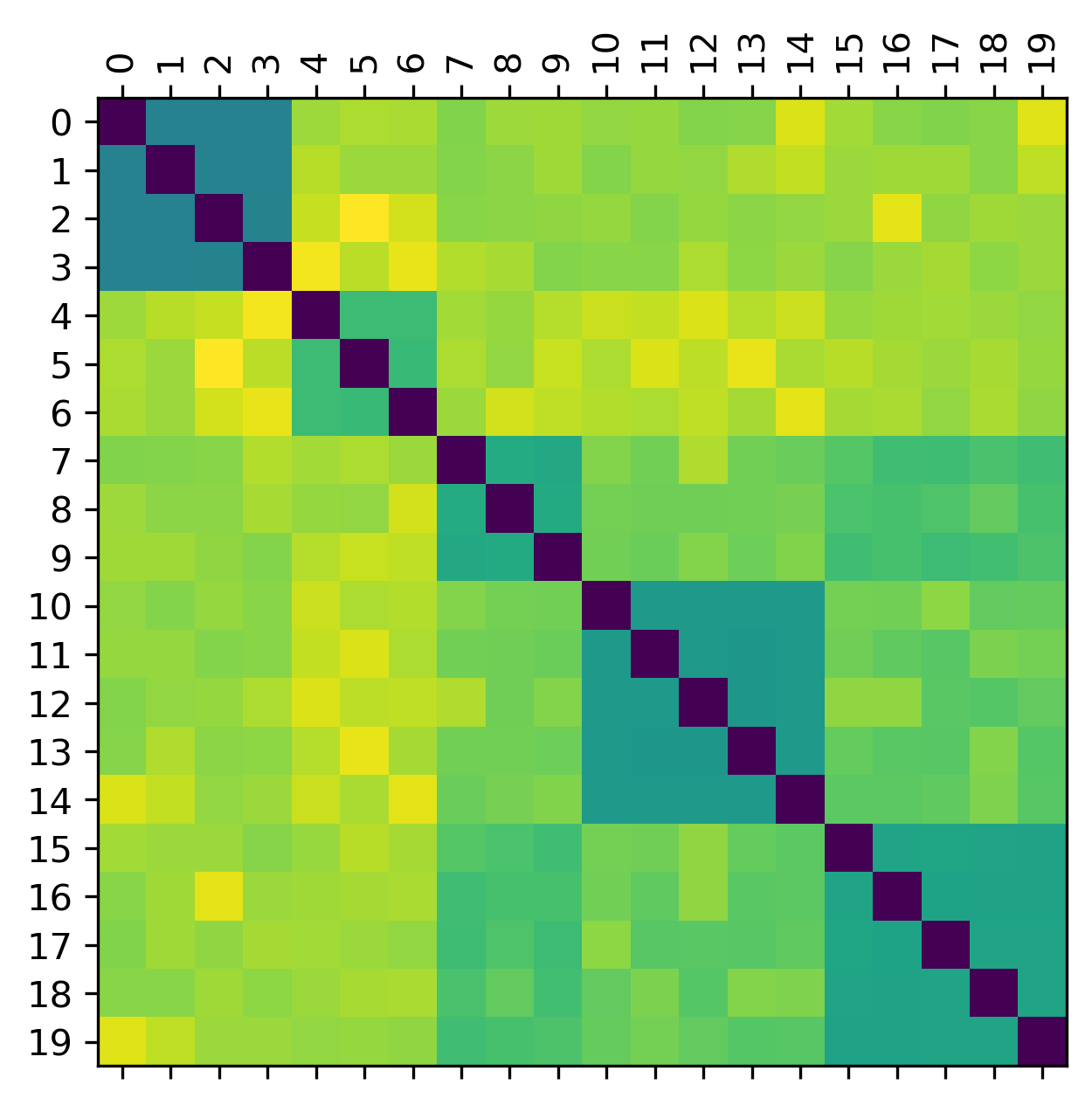}~\hfill~
~\includegraphics[width=6.5cm]{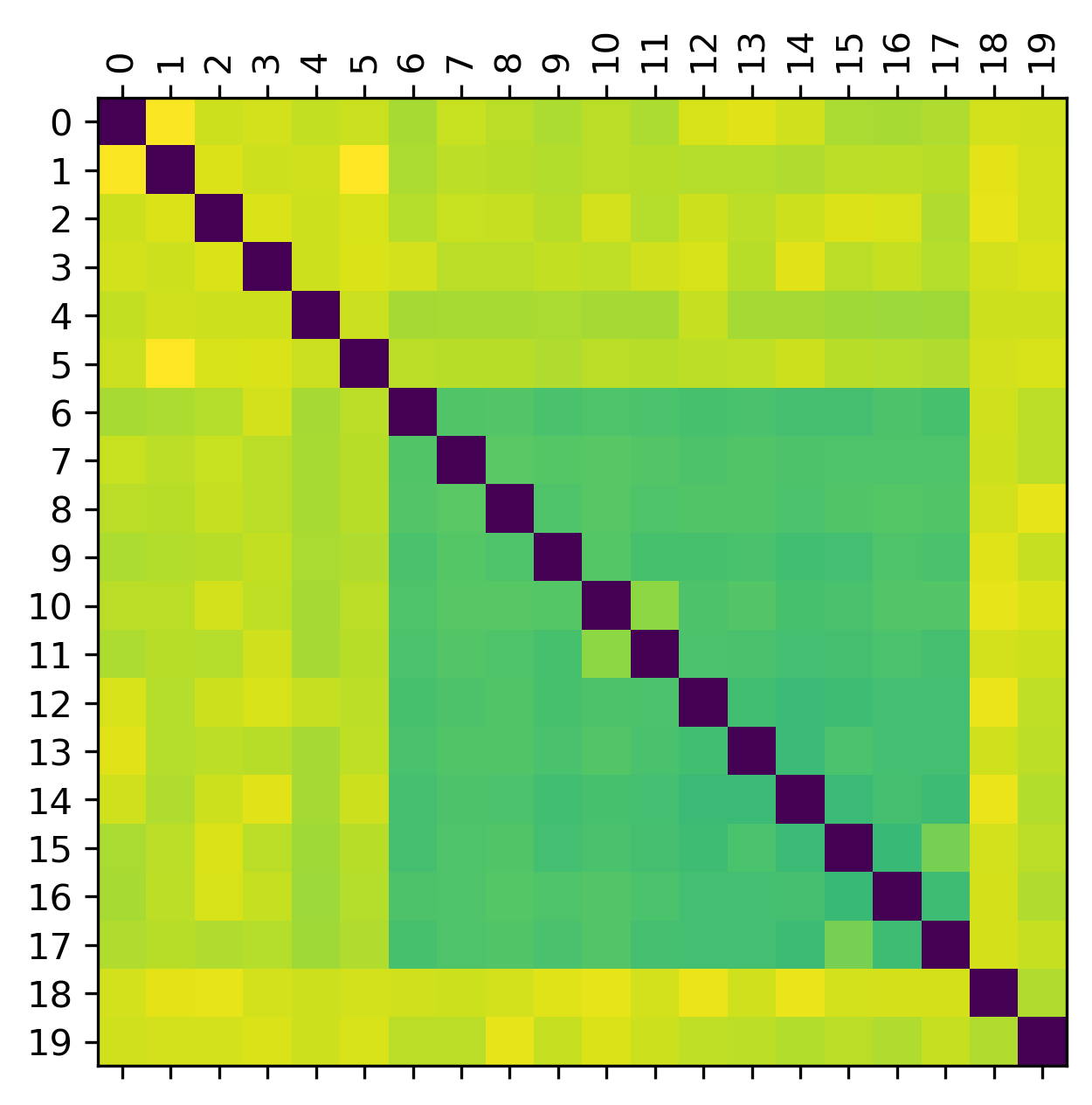}\hfill~\\
~\hfill~\includegraphics[width=6.5cm]{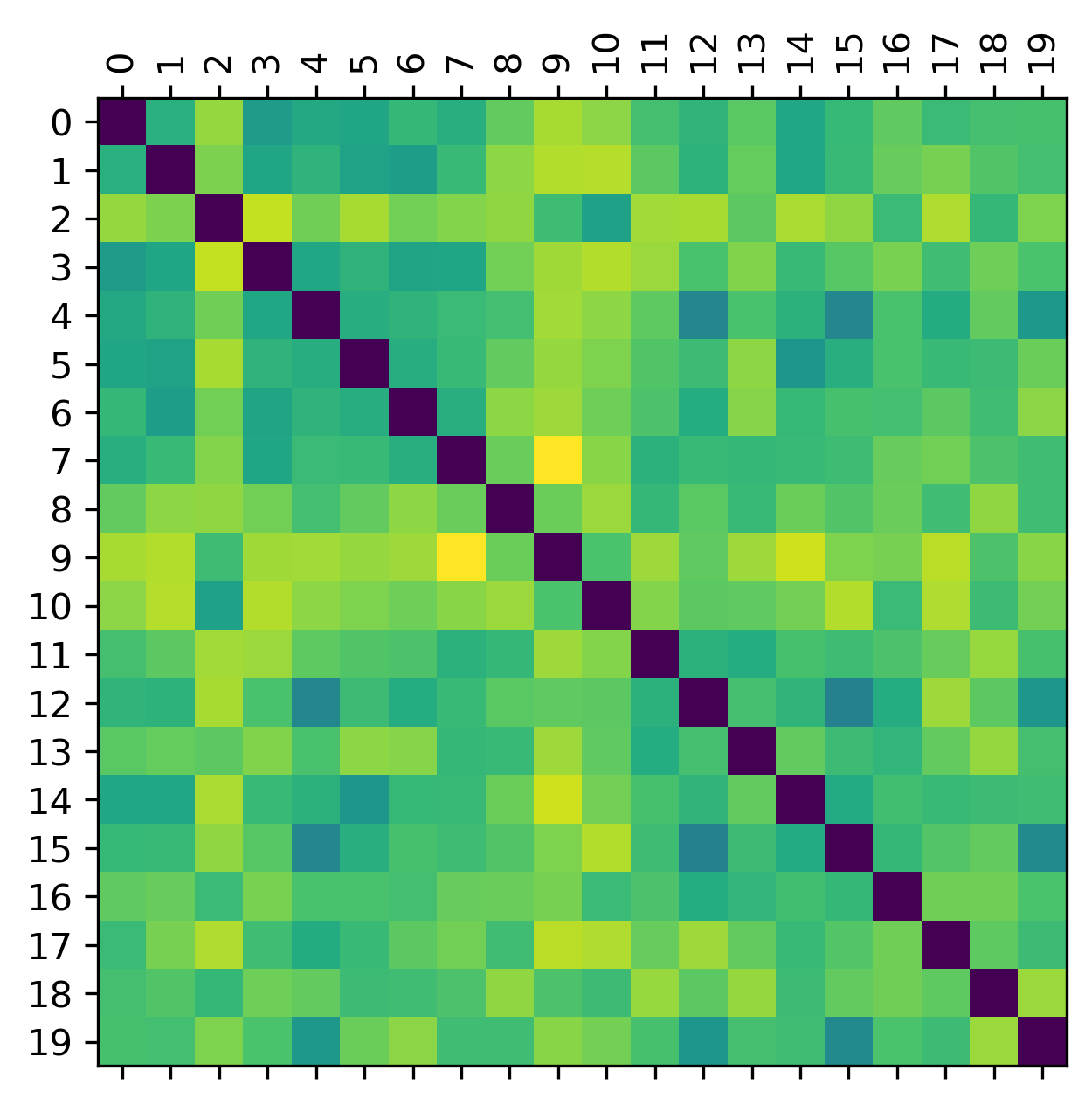}~\hfill~
~\includegraphics[width=6.5cm]{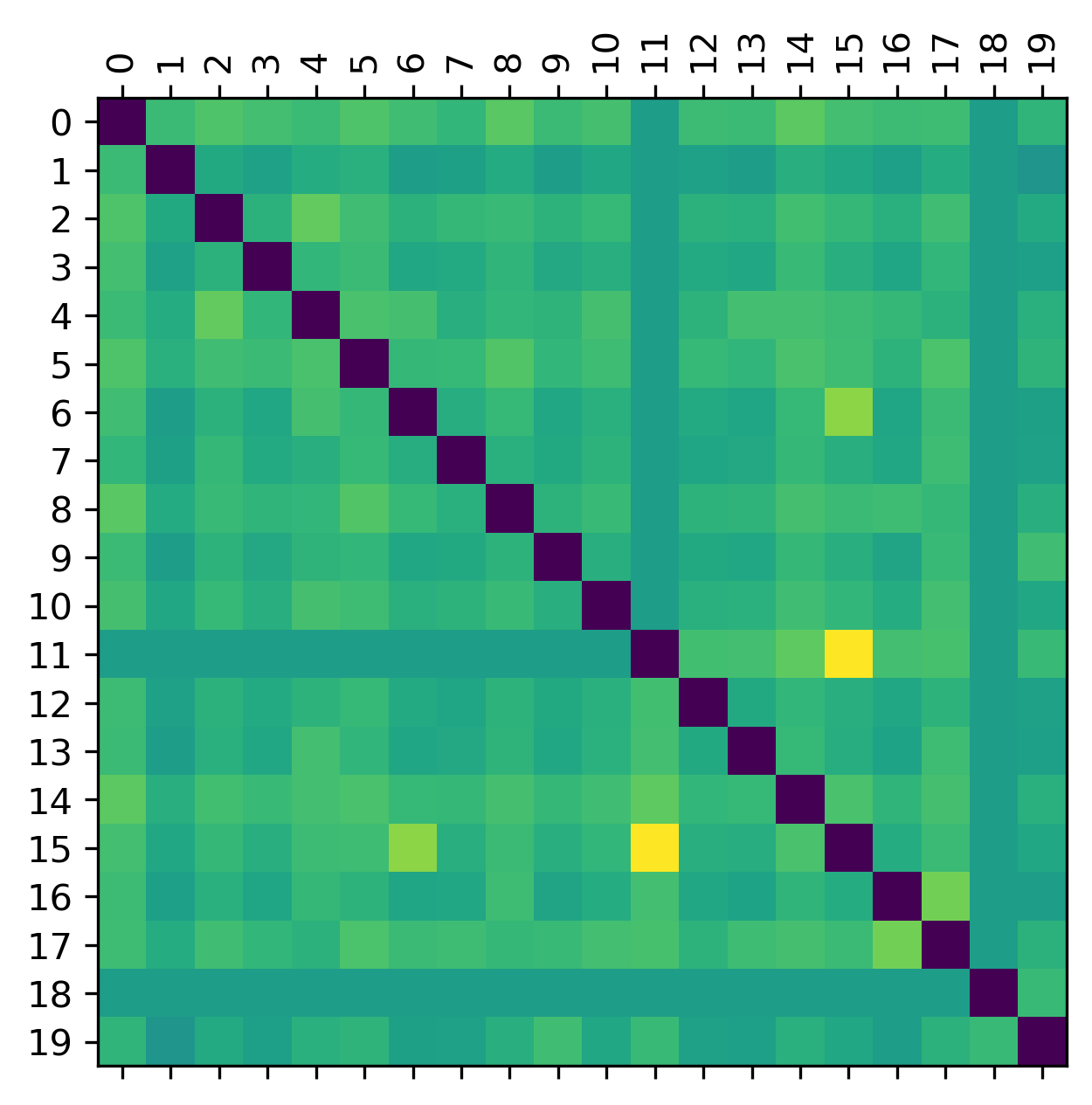}\hfill~ \\
\caption{ (left)  MNIST and (right) CIFAR10 distance matrices for $20$ clients computed using our Federated OTDD .
On the top row, we have imposed a cluster structure on client datasets while on the bottom row, there
is no specific sructure. We can note that this structure is clearly visible on the MNIST dataset
but less on CIFAR10. Eventhough, clustering clients will help improve federated learning algorithm
performances.
\label{fig:distancematrix}}
\end{figure}

We provide here more detailed results about our experiments on boosting FL methods.
\Cref*{fig:distancematrix} shows the distance matrices obtained for MNIST and CIFAR10
when the number of clients is $20$ for different structures on the clients datasets.
 We can clearly see the cluster structure on the MNIST dataset when it exists, but
 when there is no structure, the distance matrix is more uniform yet show some variations
 For CIFAR10, no clear structure is visible on the distance
  matrix as the dataset is more complex. Nonetheless, our experiments on boosting FL methods
  show that even in this case, clustering c lients can help improve the performances of federated
  learning algorithms.

Those distance matrices are the one we use as the input of the spectral clustering algorithm.
We used the spectral clustering algorithm of scikit-learn \citep{scikit-learn} with the following
setting::
\begin{itemize}
  \item we denoted as ``affinity", the setting in which the distance matrix, after rescaling, is used as affinity matrix,
  where larger values indicate greater similarity between instances. (see affinity parameter set to `precomputed' In
  scikit-learn)
  \item we denote as Sparse G. (3) and Sparse G. (5) the setting in which the distance matrix
  is interpreted as a sparse graph of distances, and construct a binary affinity matrix 
  from the  ($3$ or $5$) nearest neighbors of each instance.
   matrix is computed
\end{itemize}

\paragraph{Details on the cluster structure}  We have built this cluster structure on the client
datasets by assigning to each client one pair of
classes among the following $5$ ones : $[(0,1),(2,3),(4,5),(6,7),(8,9)]$. 
When the number of clients in equal to $10$, each cluster is composed of $2$ clients. For
a larger number of clients, each cluster is of random size with a minimum of $2$ clients.

\paragraph{Practical algorithmic details} In practice, we used the code of FedRep \cite{pmlr-v139-collins21a} for the
\emph{FedAvg}, \emph{FedRep} and \emph{FedPer} and the spectral clustering
method of scikit-learn \cite{scikit-learn}. The federated OT distance dataset has been computed
on the original data space while for CIFAR10, we have worked on the 784-dimensional code obtained from
an (untrained) randomly initialized autoencoder. We have also considered the case where the
there is no specific clustering structure on the clients as they randomly select a pair of classes
among the $10$ ones. 

\paragraph{Extra results} Performance results on federated learning are reported below for different settings.
 \Cref*{tab:clustering_mnist_grp0} and  \Cref*{tab:clustering_mnist_grp2} show the 
 results for MNIST respectively with and without client structure. 
  \Cref*{tab:clustering_cifar_grp0} and  \Cref*{tab:clustering_cifar_grp2} report
  similar results for CIFAR10.

\begin{table*}\centering
  \footnotesize
  \ra{1.1}
  \begin{tabular}{@{}rrrrrrrrrrr@{}}\toprule
  & \multicolumn{1}{c}{~} & \multicolumn{6}{c}{Clustered} \\
   \cmidrule{3-8}
  &   & \multicolumn{2}{c}{Affinity}  & \multicolumn{2}{c}{Sparse G. (3)} &\multicolumn{2}{c}{Sparse G. (5)} \\ 
    & \multicolumn{1}{c}{Vanilla} & \multicolumn{1}{c}{10} & \multicolumn{1}{c}{100} & \multicolumn{1}{c}{10} & \multicolumn{1}{c}{100} & \multicolumn{1}{c}{10} & \multicolumn{1}{c}{100} \\
  \midrule
  FedAvg\\
  10& 19.6 $\pm$ 0.9 & \textbf{99.6} $\pm$ \textbf{0.0} & \textbf{99.6} $\pm$ \textbf{0.0} & 90.5 $\pm$ 8.7 & 91.8 $\pm$ 9.6 & 84.5 $\pm$ 8.0 & 85.2 $\pm$ 5.7 \\
  20& 26.3 $\pm$ 3.8 & \textbf{99.5} $\pm$ \textbf{0.0} & \textbf{99.5} $\pm$ \textbf{0.0} & \textbf{99.5} $\pm$ \textbf{0.0} & \textbf{99.5} $\pm$ \textbf{0.0} & 91.5 $\pm$ 10.3 & 96.5 $\pm$ 6.0 \\
  40& 39.1 $\pm$ 9.0 & \textbf{99.2} $\pm$ \textbf{0.1} & \textbf{99.2} $\pm$ \textbf{0.1} & 91.1 $\pm$ 6.5 & \textbf{99.2} $\pm$ \textbf{0.1} & 94.5 $\pm$ 9.4 & \textbf{99.2} $\pm$ \textbf{0.1} \\
  100& 39.2 $\pm$ 7.7 & \textbf{98.9} $\pm$ \textbf{0.0} & \textbf{98.9} $\pm$ \textbf{0.0} & 95.9 $\pm$ 4.6 & 96.7 $\pm$ 3.8 & 98.4 $\pm$ 0.8 & \textbf{98.9} $\pm$ \textbf{0.0} \\
  FedRep\\
  10& 71.6 $\pm$ 10.5 & \textbf{99.4} $\pm$ \textbf{0.0} & \textbf{99.4} $\pm$ \textbf{0.1} & 94.3 $\pm$ 7.7 & 99.0 $\pm$ 0.5 & 95.5 $\pm$ 5.6 & 90.5 $\pm$ 6.6 \\
  20& 81.1 $\pm$ 8.1 & \textbf{99.1} $\pm$ \textbf{0.0} & \textbf{99.1} $\pm$ \textbf{0.1} & \textbf{99.1} $\pm$ \textbf{0.0} & \textbf{99.1} $\pm$ \textbf{0.0} & 98.2 $\pm$ 1.3 & 99.0 $\pm$ 0.2 \\
  40& 88.8 $\pm$ 10.4 & 98.9 $\pm$ 0.1 & 98.9 $\pm$ 0.0 & 93.3 $\pm$ 7.1 & \textbf{99.0} $\pm$ \textbf{0.1} & 96.7 $\pm$ 4.5 & \textbf{99.0} $\pm$ \textbf{0.1} \\
  100& 93.0 $\pm$ 3.9 & \textbf{98.6} $\pm$ \textbf{0.1} & \textbf{98.6} $\pm$ \textbf{0.1} & 98.4 $\pm$ 0.1 & 98.4 $\pm$ 0.1 & 98.5 $\pm$ 0.1 & 98.5 $\pm$ 0.1 \\
  FedPer\\
  10& 86.7 $\pm$ 4.3 & \textbf{99.6} $\pm$ \textbf{0.0} & \textbf{99.6} $\pm$ \textbf{0.0} & 99.5 $\pm$ 0.1 & \textbf{99.6} $\pm$ \textbf{0.1} & 98.4 $\pm$ 2.0 & 98.9 $\pm$ 1.0 \\
  20& 94.3 $\pm$ 4.3 & \textbf{99.5} $\pm$ \textbf{0.0} & \textbf{99.5} $\pm$ \textbf{0.0} & \textbf{99.5} $\pm$ \textbf{0.0} & \textbf{99.5} $\pm$ \textbf{0.0} & 99.3 $\pm$ 0.3 & \textbf{99.5} $\pm$ \textbf{0.0} \\
  40& 94.7 $\pm$ 7.6 & \textbf{99.2} $\pm$ \textbf{0.1} & \textbf{99.2} $\pm$ \textbf{0.1} & 99.1 $\pm$ 0.2 & \textbf{99.2} $\pm$ \textbf{0.1} & 97.9 $\pm$ 2.7 & \textbf{99.2} $\pm$ \textbf{0.1} \\
  100& 98.1 $\pm$ 0.1 & \textbf{98.9} $\pm$ \textbf{0.0} & \textbf{98.9} $\pm$ \textbf{0.0} & 98.8 $\pm$ 0.2 & 98.8 $\pm$ 0.1 & \textbf{98.9} $\pm$ \textbf{0.0} & \textbf{98.9} $\pm$ \textbf{0.0} \\
  \midrule Average Uplift& - & 29.8 $\pm$ 28.4& 29.8 $\pm$ 28.4& 27.2 $\pm$ 26.6& 29.0 $\pm$ 27.2& 26.7 $\pm$ 25.2& 27.6 $\pm$ 26.3\\  \bottomrule
  \end{tabular}
  \caption{\textbf{MNIST} Average performances over 5 trials of three FL algorithms: FedAvg, FedRep and
  FedPer. For each algorithm we compare the vanilla performance with the ones
  obtained after clustering the clients using the FedOTDD distance, using three different
  parameters of the spectral clustering algorithm and for a support size of $10$ and $100$.
  The number of clients varies from $10$ to $100$. For this table, datasets from clients \textbf{do
  have} a clear cluster structure \label{tab:clustering_mnist_grp0}}
  \end{table*}

\begin{table*}\centering
  \footnotesize
  \ra{1.1}
  \begin{tabular}{@{}rrrrrrrrrrr@{}}\toprule
  & \multicolumn{1}{c}{~} & \multicolumn{6}{c}{Clustered} \\
   \cmidrule{3-8}
  &   & \multicolumn{2}{c}{Affinity}  & \multicolumn{2}{c}{Sparse G. (3)} &\multicolumn{2}{c}{Sparse G. (5)} \\ 
    & \multicolumn{1}{c}{Vanilla} & \multicolumn{1}{c}{10} & \multicolumn{1}{c}{100} & \multicolumn{1}{c}{10} & \multicolumn{1}{c}{100} & \multicolumn{1}{c}{10} & \multicolumn{1}{c}{100} \\
  \midrule
FedAvg\\
10& 20.2 $\pm$ 0.6 & 81.0 $\pm$ 4.2 & \textbf{81.3} $\pm$ \textbf{4.5} & 78.0 $\pm$ 6.0 & 77.7 $\pm$ 6.6 & 71.5 $\pm$ 5.1 & 72.0 $\pm$ 6.0 \\
20& 25.1 $\pm$ 6.6 & 71.3 $\pm$ 7.3 & \textbf{72.0} $\pm$ \textbf{4.3} & 59.5 $\pm$ 3.0 & 59.5 $\pm$ 5.7 & 57.0 $\pm$ 4.4 & 60.5 $\pm$ 2.3 \\
40& 42.5 $\pm$ 10.5 & \textbf{70.8} $\pm$ \textbf{13.5} & 70.3 $\pm$ 13.3 & 60.0 $\pm$ 3.7 & 59.5 $\pm$ 10.6 & 58.1 $\pm$ 6.3 & 56.9 $\pm$ 6.1 \\
100& 52.6 $\pm$ 3.9 & 64.4 $\pm$ 9.6 & 60.4 $\pm$ 11.3 & \textbf{76.3} $\pm$ \textbf{5.4} & 68.2 $\pm$ 6.1 & 67.9 $\pm$ 6.0 & 65.4 $\pm$ 3.7 \\
FedRep\\
10& 54.3 $\pm$ 11.2 & 90.1 $\pm$ 6.7 & 90.1 $\pm$ 7.5 & 92.1 $\pm$ 4.2 & 91.8 $\pm$ 4.6 & 91.0 $\pm$ 4.4 & \textbf{94.0} $\pm$ \textbf{3.1} \\
20& 75.6 $\pm$ 9.3 & \textbf{87.5} $\pm$ \textbf{4.5} & 86.1 $\pm$ 2.6 & 81.4 $\pm$ 8.6 & 85.1 $\pm$ 6.3 & 85.3 $\pm$ 7.3 & 87.1 $\pm$ 5.5 \\
40& 78.0 $\pm$ 6.3 & \textbf{88.0} $\pm$ \textbf{4.3} & 85.4 $\pm$ 4.8 & 78.9 $\pm$ 7.9 & 74.9 $\pm$ 8.7 & 76.7 $\pm$ 5.6 & 79.6 $\pm$ 5.7 \\
100& 86.0 $\pm$ 4.8 & \textbf{91.6} $\pm$ \textbf{3.1} & 90.7 $\pm$ 3.7 & 89.1 $\pm$ 5.0 & 84.5 $\pm$ 2.9 & 86.3 $\pm$ 4.9 & 84.9 $\pm$ 3.6 \\
FedPer\\
10& 82.0 $\pm$ 10.1 & 98.4 $\pm$ 1.4 & 96.5 $\pm$ 3.5 & 96.4 $\pm$ 3.5 & 96.5 $\pm$ 3.6 & \textbf{98.5} $\pm$ \textbf{1.4} & 98.3 $\pm$ 1.3 \\
20& 90.5 $\pm$ 2.4 & 92.7 $\pm$ 1.5 & 95.4 $\pm$ 0.5 & 93.0 $\pm$ 4.3 & \textbf{96.2} $\pm$ \textbf{3.0} & 93.8 $\pm$ 2.9 & 94.5 $\pm$ 2.5 \\
40& \textbf{92.3} $\pm$ \textbf{1.3} & 90.2 $\pm$ 4.7 & 91.0 $\pm$ 4.9 & 87.7 $\pm$ 4.1 & 87.0 $\pm$ 3.7 & 89.2 $\pm$ 2.3 & 87.5 $\pm$ 5.4 \\
100& \textbf{96.6} $\pm$ \textbf{0.9} & \textbf{96.6} $\pm$ \textbf{1.6} & 96.4 $\pm$ 2.0 & 92.1 $\pm$ 3.3 & 93.0 $\pm$ 2.3 & 90.2 $\pm$ 4.9 & 86.9 $\pm$ 1.7 \\
\midrule
Average Uplift& - & 18.9 $\pm$ 18.9& 18.3 $\pm$ 19.2& 15.7 $\pm$ 18.6& 14.8 $\pm$ 18.6& 14.1 $\pm$ 17.1& 14.3 $\pm$ 18.1 \\
\bottomrule
  \end{tabular}
  \caption{\textbf{MNIST} Average performances over 5 trials of three FL algorithms: FedAvg, FedRep and
  FedPer. For each algorithm we compare the vanilla performance with the ones
  obtained after clustering the clients using the FedOTDD distance, using three different
  parameters of the spectral clustering algorithm and for a support size of $10$ and $100$.
  The number of clients varies from $10$ to $100$. For this table, datasets from clients do
  not have a clear cluster structure \label{tab:clustering_mnist_grp2}}
  \end{table*}

  \begin{table*}\centering
    \footnotesize
    \ra{1.1}
    \begin{tabular}{@{}rrrrrrrrrrr@{}}\toprule
    & \multicolumn{1}{c}{~} & \multicolumn{6}{c}{Clustered} \\
     \cmidrule{3-8}
    &   & \multicolumn{2}{c}{Affinity}  & \multicolumn{2}{c}{Sparse G. (3)} &\multicolumn{2}{c}{Sparse G. (5)} \\ 
        & \multicolumn{1}{c}{Vanilla} & \multicolumn{1}{c}{10} & \multicolumn{1}{c}{100} & \multicolumn{1}{c}{10} & \multicolumn{1}{c}{100} & \multicolumn{1}{c}{10} & \multicolumn{1}{c}{100} \\
    \midrule
    FedAvg\\
    10& 17.6 $\pm$ 1.1 & \textbf{79.1} $\pm$ \textbf{6.3} & 78.6 $\pm$ 6.0 & 61.6 $\pm$ 2.6 & 69.5 $\pm$ 5.1 & 72.2 $\pm$ 9.4 & 72.3 $\pm$ 6.0 \\
    20& 22.0 $\pm$ 2.6 & \textbf{75.1} $\pm$ \textbf{6.2} & 66.9 $\pm$ 9.1 & 42.6 $\pm$ 4.5 & 52.4 $\pm$ 17.0 & 52.2 $\pm$ 8.8 & 56.2 $\pm$ 13.6 \\
    40& 26.1 $\pm$ 7.1 & 65.9 $\pm$ 7.1 & \textbf{70.1} $\pm$ \textbf{5.7} & 36.7 $\pm$ 18.3 & 46.2 $\pm$ 15.7 & 48.8 $\pm$ 8.3 & 49.9 $\pm$ 12.1 \\
    100& 26.4 $\pm$ 4.3 & 68.0 $\pm$ 5.1 & \textbf{68.3} $\pm$ \textbf{4.7} & 37.4 $\pm$ 11.4 & 44.9 $\pm$ 13.0 & 39.8 $\pm$ 8.0 & 43.1 $\pm$ 10.4 \\
    Fedrep\\
    10& 82.4 $\pm$ 2.3 & \textbf{91.1} $\pm$ \textbf{1.2} & 90.7 $\pm$ 1.2 & 89.4 $\pm$ 0.8 & 90.3 $\pm$ 1.0 & 89.7 $\pm$ 2.3 & 90.0 $\pm$ 1.1 \\
    20& 81.8 $\pm$ 1.8 & \textbf{88.1} $\pm$ \textbf{2.0} & 85.9 $\pm$ 1.4 & 84.4 $\pm$ 0.5 & 86.0 $\pm$ 2.1 & 85.3 $\pm$ 0.5 & 86.8 $\pm$ 1.4 \\
    40& 80.3 $\pm$ 0.8 & 83.7 $\pm$ 2.0 & \textbf{86.2} $\pm$ \textbf{0.9} & 81.0 $\pm$ 2.1 & 82.3 $\pm$ 2.5 & 81.6 $\pm$ 1.7 & 82.1 $\pm$ 1.4 \\
    100& 75.0 $\pm$ 0.9 & \textbf{79.4} $\pm$ \textbf{2.3} & 78.5 $\pm$ 1.7 & 75.2 $\pm$ 2.4 & 76.3 $\pm$ 1.6 & 75.4 $\pm$ 1.5 & 76.9 $\pm$ 1.1 \\
    FedPer\\
    10& 82.1 $\pm$ 2.3 & \textbf{93.2} $\pm$ \textbf{1.1} & 93.0 $\pm$ 0.8 & 91.7 $\pm$ 0.5 & 93.0 $\pm$ 0.8 & 92.3 $\pm$ 2.0 & 92.7 $\pm$ 1.0 \\
    20& 85.4 $\pm$ 2.3 & \textbf{91.0} $\pm$ \textbf{1.9} & 89.1 $\pm$ 1.8 & 87.2 $\pm$ 0.5 & 88.7 $\pm$ 2.5 & 87.8 $\pm$ 0.9 & 89.5 $\pm$ 1.9 \\
    40& 85.9 $\pm$ 0.8 & 87.2 $\pm$ 2.2 & \textbf{89.7} $\pm$ \textbf{1.4} & 82.7 $\pm$ 2.5 & 85.4 $\pm$ 2.7 & 84.3 $\pm$ 1.9 & 84.9 $\pm$ 1.6 \\
    100& 82.2 $\pm$ 0.4 & \textbf{85.1} $\pm$ \textbf{1.8} & 83.4 $\pm$ 2.7 & 80.3 $\pm$ 2.0 & 81.3 $\pm$ 1.8 & 80.9 $\pm$ 1.7 & 82.5 $\pm$ 1.5 \\
    \midrule
    Average Uplift& - & 20.0 $\pm$ 21.3& 19.4 $\pm$ 20.8& 8.6 $\pm$ 12.5& 12.4 $\pm$ 15.1& 11.9 $\pm$ 16.0& 13.3 $\pm$ 16.1\\  \bottomrule
    \end{tabular}
    \caption{\textbf{CIFAR10} Average performances over 5 trials of three FL algorithms: FedAvg, FedRep and
    FedPer. For each algorithm we compare the vanilla performance with the ones
    obtained after clustering the clients using the FedOTDD distance, using three different
    parameters of the spectral clustering algorithm and for a support size of $10$ and $100$.
    The number of clients varies from $10$ to $100$. For this table, datasets from clients \textbf{do have} cluster structure \label{tab:clustering_cifar_grp0}}
    \end{table*}

  \begin{table*}\centering
    \footnotesize
    \ra{1.1}
    \begin{tabular}{@{}rrrrrrrrrrr@{}}\toprule
    & \multicolumn{1}{c}{~} & \multicolumn{6}{c}{Clustered} \\
     \cmidrule{3-8}
    &   & \multicolumn{2}{c}{Affinity}  & \multicolumn{2}{c}{Sparse G. (3)} &\multicolumn{2}{c}{Sparse G. (5)} \\ 
        & \multicolumn{1}{c}{Vanilla} & \multicolumn{1}{c}{10} & \multicolumn{1}{c}{100} & \multicolumn{1}{c}{10} & \multicolumn{1}{c}{100} & \multicolumn{1}{c}{10} & \multicolumn{1}{c}{100} \\
    \midrule
    FedAvg\\
    10& 18.1 $\pm$ 0.7 & 71.3 $\pm$ 7.3 & 71.0 $\pm$ 3.4 & 72.7 $\pm$ 6.2 & 72.6 $\pm$ 4.1 & \textbf{76.6} $\pm$ \textbf{2.6} & 72.4 $\pm$ 1.6 \\
    20& 23.5 $\pm$ 6.9 & \textbf{71.4} $\pm$ \textbf{9.7} & 71.2 $\pm$ 7.9 & 42.5 $\pm$ 4.7 & 47.8 $\pm$ 4.8 & 49.7 $\pm$ 4.7 & 44.4 $\pm$ 8.1 \\
    40& 26.6 $\pm$ 5.1 & \textbf{73.4} $\pm$ \textbf{15.9} & 71.1 $\pm$ 15.0 & 36.3 $\pm$ 4.5 & 30.9 $\pm$ 7.1 & 32.3 $\pm$ 11.6 & 30.3 $\pm$ 4.6 \\
    100& 27.5 $\pm$ 2.0 & \textbf{54.6} $\pm$ \textbf{10.1} & \textbf{54.6} $\pm$ \textbf{10.2} & 27.6 $\pm$ 4.1 & 29.8 $\pm$ 6.8 & 29.0 $\pm$ 3.8 & 28.3 $\pm$ 5.6 \\
    FedRep\\
    10& 83.6 $\pm$ 2.2 & 90.3 $\pm$ 3.1 & 90.3 $\pm$ 2.4 & \textbf{91.2} $\pm$ \textbf{1.6} & 91.1 $\pm$ 1.8 & 91.1 $\pm$ 2.7 & \textbf{91.2} $\pm$ \textbf{1.7} \\
    20& 85.3 $\pm$ 2.0 & 90.7 $\pm$ 2.5 & \textbf{91.5} $\pm$ \textbf{2.6} & 87.9 $\pm$ 2.0 & 88.4 $\pm$ 2.2 & 88.1 $\pm$ 1.4 & 88.6 $\pm$ 1.8 \\
    40& 84.1 $\pm$ 0.8 & \textbf{93.6} $\pm$ \textbf{2.9} & 93.3 $\pm$ 2.8 & 84.8 $\pm$ 1.7 & 84.4 $\pm$ 0.7 & 84.3 $\pm$ 0.5 & 85.3 $\pm$ 1.2 \\
    100& 77.9 $\pm$ 1.4 & 91.4 $\pm$ 2.0 & \textbf{91.6} $\pm$ \textbf{1.9} & 77.8 $\pm$ 1.7 & 78.0 $\pm$ 2.4 & 79.0 $\pm$ 1.1 & 79.4 $\pm$ 1.7 \\
    FedPer\\
    10& 83.1 $\pm$ 2.1 & 92.6 $\pm$ 2.2 & 92.7 $\pm$ 1.4 & 93.0 $\pm$ 1.4 & \textbf{93.1} $\pm$ \textbf{1.5} & 93.0 $\pm$ 2.0 & \textbf{93.1} $\pm$ \textbf{1.3} \\
    20& 88.7 $\pm$ 1.7 & 92.3 $\pm$ 1.8 & \textbf{92.7} $\pm$ \textbf{2.4} & 89.8 $\pm$ 2.0 & 90.2 $\pm$ 1.8 & 90.1 $\pm$ 1.5 & 90.0 $\pm$ 1.2 \\
    40& 88.1 $\pm$ 0.7 & \textbf{94.8} $\pm$ \textbf{2.6} & 94.6 $\pm$ 2.5 & 86.0 $\pm$ 2.3 & 86.5 $\pm$ 0.7 & 84.9 $\pm$ 3.3 & 85.7 $\pm$ 1.4 \\
    100& 85.1 $\pm$ 0.6 & 94.0 $\pm$ 1.4 & \textbf{94.1} $\pm$ \textbf{1.3} & 82.0 $\pm$ 2.4 & 82.3 $\pm$ 2.2 & 83.0 $\pm$ 1.1 & 83.6 $\pm$ 1.6 \\
    \midrule Average Uplift& - & 19.9 $\pm$ 18.0& 19.7 $\pm$ 17.5& 8.3 $\pm$ 15.2& 8.6 $\pm$ 15.4& 9.1 $\pm$ 16.6& 8.4 $\pm$ 15.1\\ \bottomrule
    \end{tabular}
    \caption{\textbf{CIFAR10} Average performances over 5 trials of three FL algorithms: FedAvg, FedRep and
    FedPer. For each algorithm we compare the vanilla performance with the ones
    obtained after clustering the clients using the FedOTDD distance, using three different
    parameters of the spectral clustering algorithm and for a support size of $10$ and $100$.
    The number of clients varies from $10$ to $100$. For this table, datasets from clients do
    not have a clear cluster structure \label{tab:clustering_cifar_grp2}}
    \end{table*}

\end{appendices}

\end{document}